\pdfoutput=1
\documentclass[11pt]{article}
\usepackage[preprint]{acl}
\usepackage{times}
\usepackage{latexsym}
\usepackage[T1]{fontenc}
\usepackage[utf8]{inputenc}
\usepackage{microtype}
\usepackage{inconsolata}
\usepackage{graphicx}
\usepackage{booktabs}
\usepackage{amsmath,amssymb}
\usepackage{xcolor}
\usepackage{fancyvrb}
\usepackage{longtable}
\usepackage{array}
\usepackage{calc}
\usepackage{tikz}
\usepackage{fontawesome5}   %
\usepackage[most]{tcolorbox}
\usepackage{enumitem}
\usepackage{placeins}   %
\usetikzlibrary{arrows.meta,fit,backgrounds}
\providecommand{\tightlist}{\setlength{\itemsep}{0pt}\setlength{\parskip}{0pt}}
\newcommand{\fitwidth}[1]{\resizebox{\ifdim\width>\linewidth\linewidth\else\width\fi}{!}{#1}}

\title{DynaDrive-Bench: Evaluating LLM Driving Agents\\ on User Experience in a Dynamic City}
\author{Jingjie Hu\textsuperscript{1,*}\quad
Lei Li\textsuperscript{2,3,*}\quad
Jiayi Su\textsuperscript{1}\quad
Runmin Cong\textsuperscript{1}\quad
Cong Zhang\textsuperscript{1}\quad
Wencheng Han\textsuperscript{1,\dag}\\[0.4em]
\normalfont
\textsuperscript{1}School of Control Science and Engineering, Shandong University, Jinan, China\\
\textsuperscript{2}Yunnao Zhilian (Shenzhen) Data Intelligence Co., Ltd., Shenzhen, China\\
\textsuperscript{3}The University of Hong Kong, Hong Kong SAR, China\\[0.3em]
\small \textsuperscript{*}Equal contribution.\quad
\textsuperscript{\dag}Corresponding author:
\texttt{hanwencheng@sdu.edu.cn}}

\begin{document}
\maketitle

\begin{abstract}
LLM driving agents manage a user\textquotesingle s driving day: when to leave, how to route, where to park or charge, and how to adapt.
Existing benchmarks judge them like navigators, checking static plans or trip-by-trip completion. Yet users care how
the whole day goes as the city changes. We introduce DynaDrive-Bench on LivingCity, a data-calibrated, minute-level
simulator over real urban road networks. Traffic, crashes, parking and charger queues, energy, and user requests evolve
independently and respond to the agent under driver-level partial observability. Each stressed bilingual day is scored
directly from execution, with no LLM judge. Completion is only the floor; the score prices time, lateness, and money
against a same-day full-information reference. Evaluating eight LLM agents shows that static plans break down in
execution, while high completion masks poor user experience. Agents finish errands but waste user time, reroute
unproductively, and most often fail on tomorrow\textquotesingle s energy. Guided by these findings, we propose SimAct, a
simulate-before-acting baseline that consistently improves performance. Together, LivingCity, DynaDrive-Bench, and
SimAct offer a reproducible testbed and direction for user-centered driving agents in changing cities.

\end{abstract}

\begin{figure}[t]
\centering
\includegraphics[width=\columnwidth]{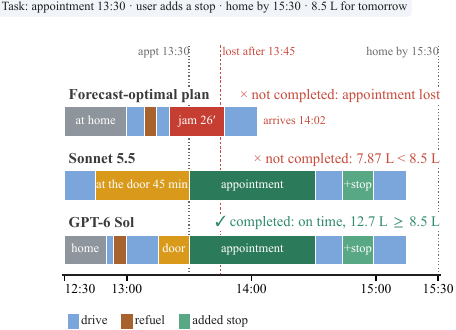}
\caption{One afternoon, three ways: of the three, only GPT-6 Sol completes the day.}
\label{fig:case}
\end{figure}

\begin{figure*}[t]
\centering
\resizebox{\textwidth}{!}{\input{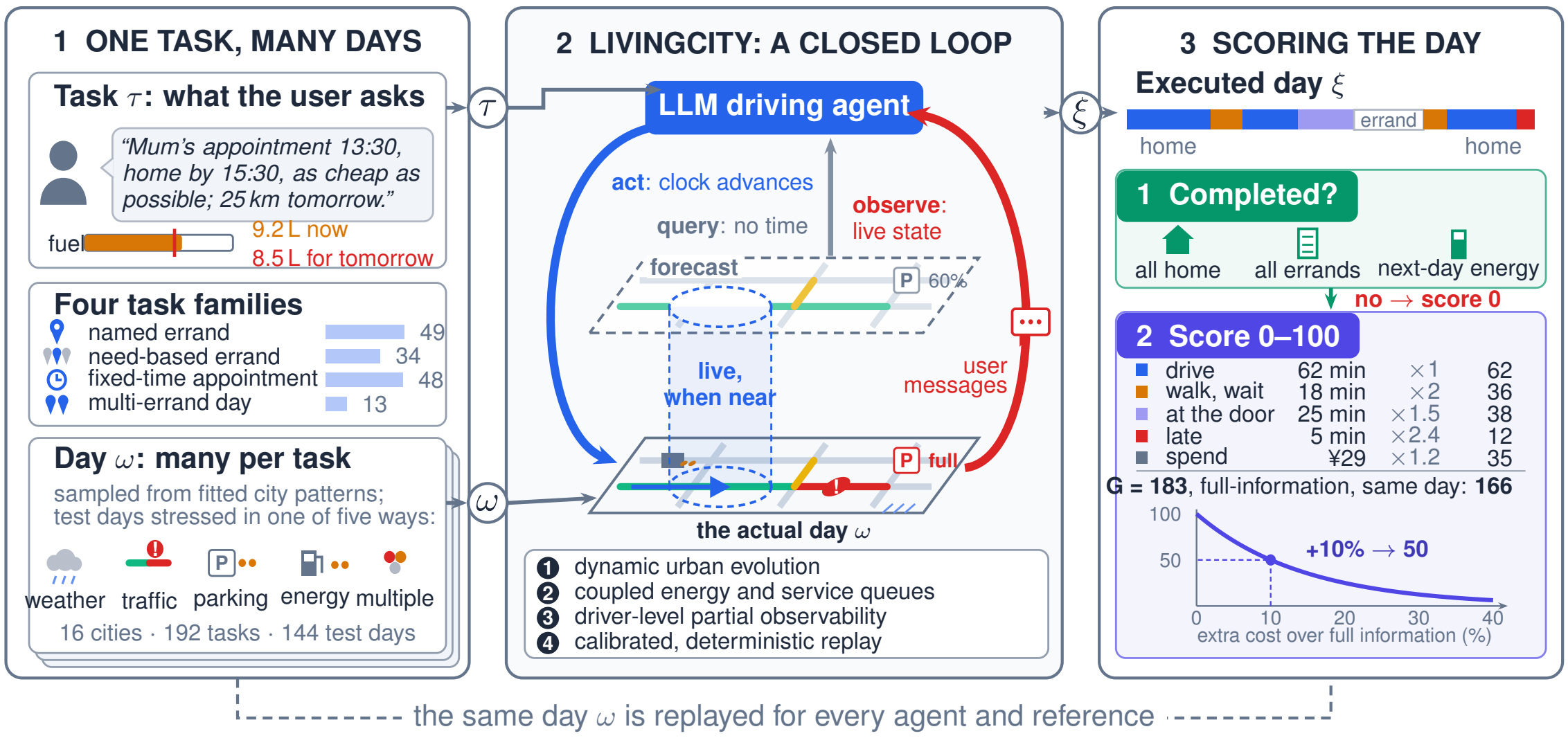}}
\caption{Pipeline overview. A task $\tau$ runs on sampled days $\omega$ in LivingCity under partial observability. The executed trace $\xi$ is checked for completion and scored 0--100 against the same-day full-information reference (halving per 10\% excess cost; values illustrative).}
\label{fig:method}
\end{figure*}

\section{Introduction}

Consider a driving agent taking a user and an elderly parent to a 13:30 hospital appointment (Fig.~\ref{fig:case}). En route, the user adds a supermarket stop; they must return home by 15:30 with fuel for tomorrow. A forecast-optimal plan hits an unforecast jam and misses the appointment. One agent leaves early to avoid lateness, but leaves the user waiting 45 minutes at the door and short of fuel at home. Another delays departure so the user waits at home, refuels en route, and finishes both errands on time. Static plans or checklists cannot separate these outcomes; only the executed day does.

Coordinating such a day is the core task of LLM driving agents operating above vehicle control \citep{xi_agentsurvey_2023, wang_agentsurvey_2023}. Yet existing benchmarks still evaluate them like navigators, scoring static itineraries \citep{jian_xie_travelplanner_2024, chen2026trip}, isolated route or station choices \citep{song_mobilitybench_2026, wu_evgeoqa_2026}, in-car tool calls \citep{kirmayr_carbench_2026}, scripted disruptions \citep{jiayu_liu_costbench_2025, karmakar_triptide_2026}, or single-drive motion control \citep{yang_vehiclearena_2026, xiaosong_jia_bench2drive_2024}. Just as autonomous driving moved from open-loop metrics to closed-loop simulation \citep{holger_caesar_nuplan_2021, daniel_dauner_navsim_2024, xiaosong_jia_bench2drive_2024}, driving agents need a city that evolves independently and answers back. Even closed-loop completion is insufficient: waiting an hour at the door or lacking tomorrow\textquotesingle s fuel ruins a completed day. We ask how well driving agents protect day-level user experience in a changing city.

Answering this requires experience-based scoring and an environment combining four properties absent from prior benchmarks (Table~\ref{tab:positioning}):

\begin{enumerate}
\def\labelenumi{\arabic{enumi}.}
\tightlist
\item
  \textbf{Dynamic urban evolution:} traffic, crashes, weather, and user requests change independently of the agent.
\item
  \textbf{Coupled energy and queues:} energy drain depends on speed and temperature, while car parks and chargers queue.
\item
  \textbf{Driver-level partial observability:} forecasts are queryable, but crashes and full lots appear only nearby.
\item
  \textbf{Calibrated, deterministic replay:} core dynamics follow real data, and every method faces the identical day.
\end{enumerate}

We build LivingCity (Fig.~\ref{fig:method}) on OpenStreetMap road networks of sixteen cities to simulate each day minute by minute. DynaDrive-Bench pairs 144 stressed English and Chinese days with deterministic execution scoring and no LLM judge. Completion is the floor: everyone returns home, all activities finish, and tomorrow\textquotesingle s energy reserve is met. Above that floor, a 0--100 score prices time, lateness, and money with transport-research weights. Each day is compared with a same-day full-information reference. Forecast-optimal fixed and rule-based online references run on the same days.

\begin{table}[!b]
\centering
\DeclareRobustCommand{\cf}{\tikz[baseline=-0.55ex]\fill (0,0) circle (2.1pt);}
\DeclareRobustCommand{\ch}{\tikz[baseline=-0.55ex]{\draw[line width=0.5pt] (0,0) circle (2.1pt); \fill (0,0) -- (90:2.1pt) arc (90:270:2.1pt) -- cycle;}}
\DeclareRobustCommand{\ce}{\tikz[baseline=-0.55ex]\draw[line width=0.5pt] (0,0) circle (2.1pt);}
\caption{Comparison with existing benchmarks (\cf\ yes, \ch\ partly, \ce\ no; criteria in Appendix~\ref{app:compare}). \emph{Closed}: closed-loop execution; \emph{Mid-trip}: mid-trip user changes; \emph{Partial}: partial observability.}
\label{tab:positioning}
\small
\setlength{\tabcolsep}{3.2pt}
\resizebox{\columnwidth}{!}{%
\begin{tabular}{@{}lcccccccc@{}}
\toprule
 & \multicolumn{3}{c}{Execution} & \multicolumn{3}{c}{City} & \multicolumn{2}{c}{Vehicle} \\
\cmidrule(lr){2-4}\cmidrule(lr){5-7}\cmidrule(lr){8-9}
 & Closed & Mid-trip & Score & Dynamic & Partial & Map & Energy & Queue \\
\midrule
TravelPlanner & \ce & \ce & \ce & \ce & \ce & \ch & \ce & \ce \\
TripTide & \ce & \ch & \ce & \ch & \ce & \ch & \ce & \ce \\
UTP-Bench & \ce & \ce & \ch & \ch & \ce & \ch & \ce & \ch \\
Travel-Sim & \ch & \ce & \cf & \cf & \ce & \cf & \ce & \ce \\
CostBench & \cf & \ch & \cf & \ch & \ce & \ce & \ce & \ce \\
MobilityBench & \ch & \ce & \ch & \ce & \ce & \cf & \ce & \ce \\
EVGeoQA & \ce & \ce & \ce & \ce & \ce & \cf & \ch & \ce \\
CAR-bench & \cf & \ce & \cf & \ce & \ch & \ce & \ch & \ce \\
VehicleArena & \cf & \ch & \cf & \cf & \ch & \cf & \ce & \ce \\
$\tau$-bench & \cf & \cf & \cf & \ce & \ch & \ce & \ce & \ce \\
Vending-Bench & \cf & \ce & \cf & \cf & \ch & \ce & \ce & \ce \\
\midrule
\textbf{DynaDrive-Bench} & \cf & \cf & \cf & \cf & \cf & \cf & \cf & \cf \\
\bottomrule
\end{tabular}}
\end{table}

Both dynamic execution and experience-based scoring overturn standard conclusions. Static checks mislead: the forecast-optimal plan completes 78.5\% of days and 38\% of hard days, and ranking by completed trip time inverts score order. Bare completion also misleads: GPT-6 Sol finishes 143 of 144 days yet scores 51, with eight agents spanning 15 to 51. Rules fail from lateness, while agents waste user time. Three findings stand out:

\begin{itemize}
\tightlist
\item
  \textbf{Agents finish errands but spend the user\textquotesingle s time.} Claude models leave at the earliest minute and wait about an hour before fixed appointments.
\item
  \textbf{Agents react, but reactions rarely help.} Counterfactual forks show crash reroutes lose time as often as they save it.
\item
  \textbf{Agents know tomorrow\textquotesingle s energy need but fail to protect it.} It is the top failure for six agents; a one-sentence reminder cuts failures by 40--68\%.
\end{itemize}

These failures stem from one bottleneck: misjudging plan outcomes. We propose SimAct, a simulate-before-acting baseline that previews candidate plans in a forecast copy of the city. Across all eight agents, it lifts mean scores from 34.6 to 48.2; ablations confirm environment rollout drives the gain.

Our contributions are fourfold: \textbf{(1) LivingCity}, a calibrated, deterministic simulator with coupled traffic, queues, and energy; \textbf{(2) DynaDrive-Bench}, 144 stressed bilingual days scored on executed user experience; \textbf{(3) a diagnostic study of eight LLM driving agents} via stability runs, counterfactual forks, and ablations; and \textbf{(4) SimAct}, a simulate-before-acting baseline validating forward consequence prediction.

\section{Related Work}\label{sec:related}

\textbf{Travel and mobility planning.} TravelPlanner \citep{jian_xie_travelplanner_2024} and its successors test whether
agents satisfy constraints over real places, personas and budgets \citep{shao2024chinatravel, zhang_deepplanning_2026,
zexuan_wang_worldtravel_2026}. Later benchmarks add change: shifting dialogue constraints \citep{chen2026trip,
yuanzhe_shen_trip-bench_2026}, disruptions \citep{karmakar_triptide_2026}, changing tool costs
\citep{jiayu_liu_costbench_2025}, or uncertain delays \citep{rao_utpbench_2026}. Map benchmarks test geospatial
reasoning and routing on real map services \citep{mahir_labib_dihan_mapeval_2024, feng2024citybench,
song_mobilitybench_2026}, and EVGeoQA \citep{wu_evgeoqa_2026} asks agents to pick a charger near an activity. Closer to
execution, TravelEval \citep{weiyi_chen_traveleval_2026} checks a finished itinerary against simulated queues, and
Travel-Sim \citep{yang_wide-horizon_2025} delegates execution to a separate traveler model. These benchmarks rely on
fixed information or score a finished plan, decoupling decisions from delays, queues and energy drain. In
DynaDrive-Bench, the agent acts through a city that changes on its own and answers back.

\textbf{Vehicles and driving.} In-car assistant benchmarks evaluate multi-turn tool calls over navigation and vehicle controls
\citep{kirmayr_carbench_2026, yang_vehicleworld_2025}. Driving benchmarks evaluate motion planning in closed loop
\citep{holger_caesar_nuplan_2021, daniel_dauner_navsim_2024, xiaosong_jia_bench2drive_2024}, and VehicleArena
\citep{yang_vehiclearena_2026} places LLM agents in 3D traffic. These benchmarks score one session or one drive. Our
driving agent works above vehicle control and is judged over a user\textquotesingle s day of driving, from home to home.

\textbf{Outcome-based evaluation.} \(\tau\)-bench \citep{yao_taubench_2024}, \(\tau^2\)-bench \citep{barres_tau2bench_2026} and
Vending-Bench \citep{backlund_vendingbench_2025} score agents by the final state of the world and by consistency across
runs. \citet{zhu_abc_2025} warn that flaws in task setup can distort rankings, and agents can exploit how they are
scored \citep{bondarenko_specgaming_2025, thaman_rhb_2026}. These benchmarks ask whether the task succeeded, ignoring
what the execution costs the user. We treat completion only as the floor and score how the executed day went for the
user, against a same-day full-information reference. Transport research supports this choice. Travelers value
waiting, walking and lateness differently from driving time \citep{wardman_multipliers_2026, small_scheduling_1982,
shoup_cruising_2006}, and routing is coupled with charging and parking \citep{montoya_electric_2017,
tobias_arndt_probabilistic_2016}. Table~\ref{tab:positioning} summarizes these differences.

\textbf{Planning before acting.} Some agent frameworks write an explicit plan before or alongside ReAct-style acting
\citep{erdogan_planact_2025, rawat_preact_2025}. World-model agents go further: they simulate candidate futures before
committing, with an LLM as the world model \citep{deng_simura_2025, yu_dynamind_2026}. Unlike LLM world models, our SimAct
baseline delegates rollout to a forecast copy of LivingCity rather than the model itself.

\section{LivingCity and DynaDrive-Bench}

Figure~\ref{fig:method} summarizes our pipeline: each task \(\tau\) runs on a sampled city day \(\omega\) in LivingCity, producing a trace \(\xi\) scored directly on user experience.

\subsection{The LivingCity Dynamic Simulator}

LivingCity simulates a city day \(\omega\) at minute resolution as a closed loop. The city evolves whether the agent moves or waits, and each action shapes what follows. State \(s_t\) tracks the person, vehicle, companion, energy, completed activities, spend, and city traffic and facilities. Each turn, the agent queries \(o_t = O(s_t)\) at zero simulated time. It then executes a physical action, which advances the clock deterministically given \(\omega\): \(s_{t'} = T(s_t, a_t; \omega)\).

\textbf{Coupled urban dynamics.} Each road segment sits in a 2-km grid cell with a daily speed profile per road class and day-to-day variation. Speeds drop under adverse weather \citep{fhwa_weather_2006} and unforecast crashes lasting tens of minutes \citep{shen_incident_2024}. Energy drain depends on speed, idling, and ambient temperature. Car parks and chargers act as finite-capacity queues with time-of-day pricing, spilling overflow to nearby lots. At busy venues, agents must weigh queueing against reserving ahead or parking further and walking.

\textbf{Driver-level observability and interruptions.} Forecasts are always queryable, but live observations match what a driver sees (Fig.~\ref{fig:method}). Crashes appear only within roughly 1.5 km. Full lots appear only at the entrance unless they publish a live feed; only one in five does, with delay and noise. Each turn, the agent submits one action---driving, parking, queueing, or charging---with a one-sentence reason. LivingCity interrupts an ongoing action upon a crash ahead, projected lateness, a filling destination lot, low energy, or a mid-trip user message. The agent then continues or replans.

\textbf{Calibration and deterministic replay.} Congestion curves are fitted to TomTom\footnote{\url{https://www.tomtom.com/traffic-index/}} and Baidu\footnote{\url{https://jiaotong.baidu.com/}} indices. Across 537 AMap\footnote{\url{https://lbs.amap.com/}} routes in the Chinese cities, the median length ratio is 0.985, with 83\% of travel times within \(\pm 25\%\). Parking and charger queues match peak occupancy in Birmingham and Shenzhen \citep{stolfi_birmingham_2017, li_urbanev_2026}, and charger fault rates follow \citet{rempel_dcfc_2022}. Remaining parameters are set by design (Appendix~\ref{app:sim}; tools in Appendix~\ref{app:bench}). Deterministic \(T\) ensures identical days across methods and exact counterfactual forks at any decision point.

\subsection{Tasks, Stressed Days and Dataset}

\textbf{Tasks and days.} Each input pairs a morning task \(\tau\) with a realized day \(\omega\). A \emph{task} \(\tau\) gives an English or Chinese request with activities, earliest departure, target arrivals, tomorrow\textquotesingle s commute distance, home, companion, notices, and starting energy. A \emph{day} \(\omega\) specifies minute-level congestion, facility arrivals, crashes, weather, charger faults, and mid-trip user messages. The agent sees \(\tau\) upfront and uncovers \(\omega\) during execution. Like a leave-by alert, the agent proposes departure time after the user\textquotesingle s earliest availability, and the user accepts.

\textbf{Task families and stressed regimes.} Requests cover four families: \emph{named errands} (a fixed venue), \emph{need-based errands} (venue choice among candidates), \emph{fixed-time appointments} (lost after a short grace period, penalizing both lateness and early door waits), and \emph{multi-errand days} (two coupled stops). Every request includes tomorrow\textquotesingle s commute. Each day draws from five stressed regimes: weather, traffic, parking, energy, or combined. Days also mix in mid-trip messages, road closures, events, rain, and companions with walking limits.

\textbf{Grounded construction and statistics.} Real places are sampled from \(24 \times 24\) km maps via AMap (ten Chinese cities) and OpenStreetMap (London, New York, Los Angeles, Singapore, Sydney, Toronto). Tasks place one or two activities 1.2--14 km from home during peak hours, with tight energy and 10--60 minutes of initial slack. Sampled days enter the test set only if a reference solver completes them (143 by the full-information search and 1 by the fixed plan; Appendix~\ref{app:scoring}). The full package has 192 tasks across sixteen cities and 1,496 inputs (four development cities). The frozen test set has 144 stressed days across twelve cities: 49 named errands, 34 need-based, 48 appointments, and 13 multi-errand days. Half use electric and half combustion vehicles. Of these days, 51 include mid-trip messages and 71 include road-closing crashes. Fifty are \emph{hard}: the unchanged forecast-optimal plan fails or ends below tomorrow\textquotesingle s reserve.

\subsection{Scoring the User\textquotesingle s Experience}

Executing agent \(\pi\) on \((\tau, \omega)\) yields trace \(\xi = \mathrm{Exec}(\pi;\tau,\omega) = ((t_k, a_k, s_k))_{k=0}^{K}\), recording positions \(\ell_p, \ell_v, \ell_c\) (person, vehicle, any companion), energy \(e(t)\), completed activities \(D(\xi)\), and arrival times \(\alpha_j(\xi)\). Scoring is deterministic from \(\xi\) with no LLM judge.

\textbf{Experience of the executed day.} Between leaving home (\(t_0\)) and returning (\(t_K\)), each minute lies in state \(\sigma_\xi(t) \in \mathcal{C}\) = \{drive, walk, find/queue, wait for energy, wait at door, other wait, activity\}. Trip time \(\mathrm{TT}(\xi) = \sum_{c} T_c(\xi)\) splits into state minutes \(T_c(\xi)\) weighted differently by travelers \citep{wardman_multipliers_2026, small_scheduling_1982}. We also track walking distance \(W(\xi)\) \citep{shoup_cruising_2006} and spend \(P(\xi)\). Total lateness \(\Lambda(\xi) = \sum_j \max\big(0, \alpha_j(\xi) - \bar t_j\big)\) sums over all target times \(\bar t_j\), including the return home and mid-trip updates. An appointment with grace \(g_j\) is kept if \(\alpha_j(\xi) \le \bar t_j + g_j\).

\textbf{Completion: the floor.} Required activities \(R(\xi) = \big(R_0(\tau) \cup R^{+}(M(\xi))\big) \setminus R^{-}(M(\xi))\) combine initial requests with mid-trip updates \(M(\xi)\). Tomorrow\textquotesingle s commute \(d\) sets the home reserve:
\begin{equation}\label{eq:reserve}
\hat e(\tau) = \tfrac12\big\lceil 2\,\big(d\,\kappa\,f(\theta) + e_{\min}\big)\big\rceil,
\end{equation}
using nominal consumption \(\kappa\), temperature factor \(f(\theta)\), and minimum buffer \(e_{\min}\). A day is \emph{completed} if everyone returns home (\(h\)), all required activities finish, and home energy meets \(\hat e(\tau)\) within \(\epsilon = 0.1\,e_{\min}\):
\begin{equation}\label{eq:complete}
\begin{split}
C(\xi) = {}& \mathbf{1}\big[\ell_p(t_K) = \ell_v(t_K) = \ell_c(t_K) = h\big]\\
&\cdot \mathbf{1}\big[R(\xi) \subseteq D(\xi)\big]\\
&\cdot \mathbf{1}\big[e(t_K) \ge \hat e(\tau) - \epsilon\big].
\end{split}
\end{equation}
Completion enforces hard deadlines (closing hours and grace periods); preferred times enter through the score.

\textbf{Same-day references.} Each day runs three references on \(\omega\). The \emph{fixed plan} follows the forecast-optimal schedule without live queries. The rule-based \emph{online replanner} queries live data, reroutes, and handles messages. The \emph{full-information reference} \(\xi^*\) knows \(\omega\) in advance and searches a fixed plan family for the minimum generalized cost. It is a strong reference, not a global optimum: it searches departures in 10-minute steps and takes the quickest route on each leg (Appendix~\ref{app:scoring}).

\textbf{Generalized cost.} Following transport economics, we price a day in equivalent driving minutes:
\begin{equation}\label{eq:gcost}
G(\xi) = \sum_{c} w_c\,T_c(\xi) + w_\Lambda\,\Lambda(\xi) + w_P\,\tilde P(\xi).
\end{equation}
Weights follow transport meta-analyses \citep{wardman_multipliers_2026, wardman_vtts_2023}. Walking, queueing, and waiting use the conventional \(2\times\) factor (606-study meta-analysis mean: 1.6--1.7). Lateness counts \(2.4\times\) (within the 2.2--3.0 urban car range), and door waiting counts \(1.5\times\) (between early arrival near \(1\times\) and active waiting). Net spend \(\tilde P(\xi)\) is today\textquotesingle s payments plus the tank energy drawn, at the median map price. It converts to minutes at personal-trip values of time: 50 yuan/h in Chinese cities and 128--176 yuan/h elsewhere. Constant activity time is excluded.

\textbf{Score.} A completed day costing no more than \(\xi^*\) earns 100. Every extra \(H = 10\%\) above \(G(\xi^*)\) halves the score, and an incomplete day earns 0:
\begin{equation}\label{eq:score}
s(\xi) = 100\; C(\xi)\; 2^{-\max(0,\;G(\xi)/G(\xi^{*})-1)/H}.
\end{equation}
Travelers judge travel-time differences proportionally and ignore small gaps \citep{mahmassani_brue_1987, di_indifference_2017}. In route-choice experiments, most switch only for a \(\ge 30\%\) gain \citep{gonzalez_ramirez_routechoice_2021}, so \(H = 10\%\) is strict. Because \(\xi^*\) is fixed per day, scores stay comparable as new agents are evaluated. We also report completion rate \(\Pr_{\mathcal{D}}[C(\xi)=1]\), failure causes, consistency \(\mathrm{Pass}^k(\pi) = \mathbb{E}_{\mathcal{D}}[\prod_{r=1}^{k} C(\xi^{(r)})]\) \citep{yao_taubench_2024}, and excess cost \(\Delta(\pi) = \mathbb{E}_{\mathcal{D}}[G(\xi) - G(\xi^{*}) \mid C(\xi) = C(\xi^{*}) = 1]\). Because prompts omit these weights, the score tests whether agents protect user experience without knowing the yardstick (sensitivity in Appendix~\ref{app:scoring}).

\section{Experiments}

\subsection{Setup}

We evaluate eight LLM agents across five families in the same ReAct loop \citep{yao_react_2023}: GPT-6 Sol, GPT-5.6 Terra, GPT-6 Luna, Claude Opus 5.5, Claude Sonnet 5.5, GLM-5.3-Flash, DeepSeek-V4.1-Flash, and MiMo-V2.6-Flash (shortened to Opus 5.5, Sonnet 5.5, GLM-5.3F, DS-V4.1F, MiMo-V2.6F). All agents share identical prompts, tool schemas, compact JSON outputs, and per-day budgets (60 actions, 100 rounds, 250 tool calls, 2,048 output tokens/turn, parallel queries). We run all eight agents once on the 144 test days and four agents twice more on a 60-day stability subset. All eight also run under SimAct (Section~\ref{sec:framework}). The three same-day references run on every day.

\subsection{Main Results}

\textbf{Agents finish their days, but few days go well for the user.} Six agents leave at most 19 of 144 days unfinished, yet scores span only 15--51 (Table~\ref{tab:main}). Each finished day loses points for extra time and spend above the full-information reference. GPT-6 Sol finishes 143 days but earns 100 on only 34; its median finished day costs 11\% more than the reference. It leads all agents across half-lives from 5\% to 40\% and under a 5-minute absolute half-life.

\begin{table*}[t]
\centering
\small
\setlength{\tabcolsep}{4pt}
\caption{Main results on 144 test days (94 calm, 50 hard), ordered by score (Eq.~\ref{eq:score}).
Drop is calm minus hard before rounding. $\Delta$ is excess generalized minutes on days completed by both the method and the full-information reference.
Parentheses show SimAct gains over ReAct (Score) and next-day energy failures (Not completed).}
\label{tab:main}
\fitwidth{%
\begin{tabular}{lrrrrrrrrrrrrr}
\toprule
 & \multicolumn{6}{c}{Score (0--100) and completion} & \multicolumn{2}{c}{On time (\%)} & \multicolumn{4}{c}{Cost on completed days} & Agent \\
\cmidrule(lr){2-7}\cmidrule(lr){8-9}\cmidrule(lr){10-13}\cmidrule(lr){14-14}
Method & \shortstack[r]{Score\\{}[95\% CI]} & \shortstack[r]{Com-\\{}pleted (\%)} & Calm & Hard & Drop & \shortstack[r]{Not com-\\{}pleted} & Appts. & Home & \shortstack[r]{$\Delta$\\{}(min)} & \shortstack[r]{Early wait\\{}(min)} & \shortstack[r]{Walk\\{}(m)} & \shortstack[r]{Spend\\{}(CNY)} & \shortstack[r]{Tokens/day\\{}(k, in / out)} \\
\midrule
GPT-6 Sol & \textbf{50.7} [45, 57] & \textbf{99.3} & \textbf{55.9} & \textbf{41.1} & 14.8 & 1 & 92.9 & 88.9 & \textbf{+25.0} & 7.2 & 354 & 70.8 & 184 / 1.5 \\
GLM-5.3F & 40.2 [34, 47] & 95.8 & 44.5 & 32.2 & 12.3 & 6 (2) & 72.3 & 84.7 & +48.7 & 12.3 & 373 & 81.7 & 326 / 8.4 \\
GPT-5.6 Terra & 39.3 [33, 46] & 95.8 & 42.6 & 33.1 & 9.5 & 6 (5) & 89.3 & 85.4 & +46.5 & 15.9 & 357 & 74.9 & 144 / 1.7 \\
Opus 5.5 & 38.3 [32, 46] & 93.1 & 44.1 & 27.4 & 16.7 & 10 (8) & 94.6 & 86.8 & +65.2 & 25.8 & 373 & 81.3 & 200 / 2.1 \\
Sonnet 5.5 & 36.4 [30, 44] & 91.0 & 42.7 & 24.4 & 18.3 & 13 (10) & 96.4 & 86.1 & +68.7 & 26.1 & 331 & 77.8 & 196 / 2.1 \\
DS-V4.1F & 29.6 [23, 36] & 86.8 & 33.1 & 22.9 & 10.2 & 19 (12) & 71.4 & 81.9 & +64.8 & 13.2 & 339 & 89.0 & 317 / 2.4 \\
GPT-6 Luna & 27.5 [21, 34] & 50.7 & 29.8 & 23.2 & 6.5 & 71 (41) & 71.9 & 77.1 & +34.5 & 18.8 & 293 & 52.4 & 118 / 0.7 \\
MiMo-V2.6F & 15.1 [11, 20] & 79.9 & 16.2 & 13.1 & 3.1 & 29 (16) & 62.5 & 75.0 & +105.1 & 13.9 & 365 & 131.2 & 544 / 1.7 \\
\midrule
\multicolumn{14}{l}{\textit{The same agents with SimAct (Section~\ref{sec:framework})}} \\
Opus 5.5 & 62.5 (+24.2) [56, 69] & 98.6 & 64.0 & 59.6 & 4.4 & 2 & 76.8 & 89.6 & +19.9 & 4.6 & 348 & 68.1 & 616 / 24.9 \\
GPT-6 Sol & 59.7 (+8.9) [54, 66] & 97.2 & 61.7 & 55.8 & 5.9 & 4 & 75.0 & 84.7 & +19.5 & 4.5 & 347 & 65.8 & 331 / 2.6 \\
GLM-5.3F & 53.5 (+13.3) [47, 60] & 95.8 & 62.0 & 37.4 & 24.7 & 6 (1) & 71.4 & 87.5 & +29.4 & 4.8 & 333 & 83.1 & 541 / 12.8 \\
GPT-5.6 Terra & 53.1 (+13.9) [47, 60] & 97.2 & 58.2 & 43.6 & 14.7 & 4 (1) & 76.8 & 84.7 & +27.6 & 6.5 & 333 & 62.2 & 266 / 2.7 \\
Sonnet 5.5 & 44.6 (+8.2) [38, 51] & 93.8 & 51.0 & 32.6 & 18.4 & 9 (1) & 71.4 & 81.2 & +49.1 & 10.4 & 394 & 94.8 & 645 / 15.0 \\
GPT-6 Luna & 44.3 (+16.8) [38, 51] & 79.9 & 45.1 & 42.7 & 2.4 & 29 (12) & 70.7 & 84.7 & +30.9 & 7.3 & 319 & 64.3 & 283 / 1.5 \\
DS-V4.1F & 37.5 (+7.9) [31, 44] & 95.8 & 44.5 & 24.2 & 20.3 & 6 (1) & 82.1 & 81.9 & +49.7 & 10.2 & 338 & 86.9 & 610 / 4.2 \\
MiMo-V2.6F & 30.9 (+15.8) [25, 37] & 93.1 & 35.3 & 22.5 & 12.7 & 10 (1) & 78.6 & 80.6 & +65.5 & 9.8 & 349 & 117.2 & 908 / 4.8 \\
\midrule
Fixed plan & 36.5 & 78.5 & 45.1 & 20.3 & 24.8 & 31 (6) & 33.0 & 67.4 & +29.4 & 1.1 & 351 & 64.0 & -- \\
Online replanner & 41.1 & 86.8 & 50.1 & 24.0 & 26.1 & 19 & 38.4 & 73.6 & +28.0 & 1.1 & 387 & 72.6 & -- \\
Full-information reference & 100.0 & 100.0 & 100.0 & 100.0 & 0.0 & 0 & 66.1 & 93.1 & 0 & 1.4 & 372 & 64.9 & -- \\
\bottomrule
\end{tabular}}
\end{table*}

\textbf{Rules do well on calm days and collapse when the city changes.} On the 94 calm days, the online replanner scores 50 (second to GPT-6 Sol\textquotesingle s 56), and the fixed plan\textquotesingle s 45 beats seven agents. On the 50 hard days, however, both rule-based scores drop about 25 points to 20--24, leaving 18--31 days unfinished. The five strongest agents drop only 10--18 points (scoring 24--41) and leave just 1--7 hard days unfinished.

\textbf{A forecast-optimal plan fails in the realized city.} Statically, the fixed plan completes all 144 days. Executed, it completes 78.5\% overall and 38\% of hard days (only when its energy shortfall stays within tolerance). Of its 31 failures, 21 miss appointments, 6 miss next-day energy, 3 leave activities undone, and 1 misses a mid-trip request. Even the online replanner loses 16 appointments.

\begin{figure}[t]
\centering
\includegraphics[width=\columnwidth]{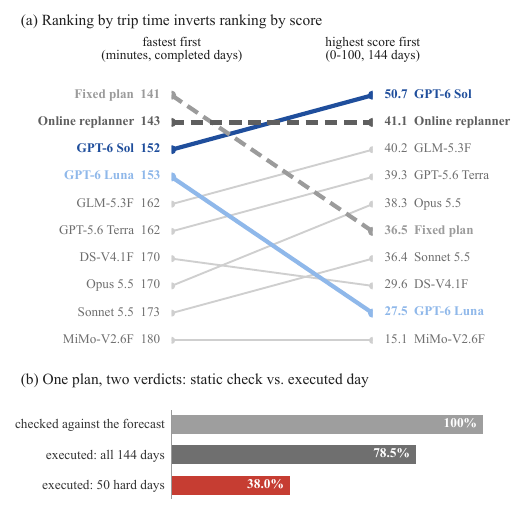}
\caption{Execution overturns static evaluation. (a) Ranking by completed trip time inverts the score order. (b) The forecast-optimal plan passes all static checks but completes 78.5\% of executed days (38\% of hard days).}
\label{fig:rankflip}
\end{figure}

\textbf{Judging trip time or finished errands alone misleads.} Ranked by mean trip time on completed days, the fixed plan is first and GPT-6 Luna fourth. By score they fall to sixth and ninth (Fig.~\ref{fig:rankflip}). Dropping their 31 and 71 failed days makes them look fast. Counting completion errs the other way by hiding wasted time; the score captures both.

\textbf{Rankings are stable across runs.} Over three runs on the 60-day subset, scores vary by at most 6 points, and habits recur every run. Multi-run consistency (\(\mathrm{Pass}^3\)) widens gaps further: GPT-6 Sol reaches 96.7\% versus 43.3\% for GPT-6 Luna.

\subsection{Where the User\textquotesingle s Time Goes}

\textbf{Agents finish errands but spend the user\textquotesingle s time.} No agent loses an appointment after the grace period. GPT-6 Sol and both Claude models arrive on time at 93--96\% of fixed-time appointments, versus 33--38\% for the rules. Rules trust forecasts and arrive late on 41--42\% of completed days. Agents instead pay for punctuality with the user\textquotesingle s time. The Claude models, GPT-5.6 Terra, and GPT-6 Luna leave at the earliest allowed minute on 87--100\% of days. Before fixed-time appointments, Claude models make users wait about an hour at the door, versus 3 minutes for the full-information reference. Across all completed days they wait 26 minutes and add \(+65.2\) to \(+68.7\) generalized minutes. DS-V4.1F and MiMo-V2.6F lose time on the road instead, accumulating charging queues, detours, and extra spend (\(+105.1\) generalized minutes and 131.2 CNY for MiMo-V2.6F). GPT-6 Sol delays departure by 16 minutes on average, keeping excess cost to \(+25.0\) minutes and 70.8 CNY.

\begin{figure*}[t]
\centering
\begin{minipage}[b]{0.49\textwidth}\centering
\includegraphics[width=\linewidth]{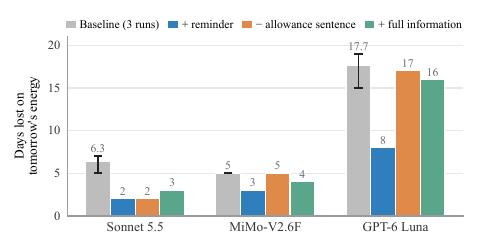}\\[-1pt]
{\small (a) Ablations on the 60-input subset}
\end{minipage}\hfill
\begin{minipage}[b]{0.49\textwidth}\centering
\includegraphics[width=\linewidth]{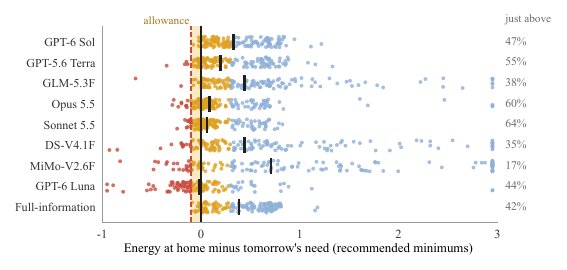}\\[-1pt]
{\small (b) Energy left for tomorrow}
\end{minipage}
\caption{Tomorrow\textquotesingle s energy: what removes the incomplete days (a) and how close to the line agents arrive home (b).}
\label{fig:energy}
\end{figure*}

\subsection{Reacting to Alerts}

\textbf{Agents react, but their reactions rarely improve the day.} At a crash alert, every agent except GPT-6 Luna reroutes in 74--95\% of cases (57 of 60 for GPT-6 Sol). The online replanner reroutes 25 of 72 times, and GPT-6 Luna 8 of 44. Because LivingCity is deterministic, we fork each alert on the identical day. One branch stays the course and the other takes the chosen detour; we compare their arrival at the destination (Appendix~\ref{app:alerts}). Rerouting is faster about as often as slower (23 versus 23 for GPT-6 Sol), and every agent\textquotesingle s median gain is slightly negative. Mean gains stay positive, except for MiMo-V2.6F, only because a few detours avoid full road closures. Forks at full car parks compare queueing with switching at the venue door. Agents pick the better option in 33--76\% of comparable cases, except GPT-6 Luna (13 of 13). They seldom reserve or call ahead.

\subsection{Tomorrow\textquotesingle s Energy}

\textbf{Agents know tomorrow\textquotesingle s energy need but fail to protect it.} Unlike rules, no agent fails by losing an appointment. Instead, next-day commute energy is the top failure mode for six agents. It causes half of the failures of GPT-6 Luna and MiMo-V2.6F, and 8 of 10 for Opus 5.5. Across 1,152 runs, 155 fail: 94 \emph{only} on next-day energy, 27 on undone activities, 25 abandoned (23 by GPT-6 Luna), 3 stranded on the road, and 6 from other causes. Three 60-day ablations isolate why (Fig.~\ref{fig:energy}a).

\emph{Not seeing?} Exact city-wide live information barely changes next-day failures for MiMo-V2.6F (5 to 4) and GPT-6 Luna (17.7 to 16). Sonnet 5.5 drops from 6.3 to 3. Observability is not the main bottleneck. \emph{Not knowing?} Every next-day failure of GPT-5.6 Terra, GLM-5.3F, and Claude explicitly notes tomorrow\textquotesingle s commute in its trace. Yet a one-sentence reserve reminder cuts next-day failures by 40--68\% across all three tested agents, showing that known constraints decay over long horizons. \emph{Not committing.} Claude models cite the 10\% shortfall allowance, aim just above it, and repeatedly land just below. Their median home margin is 0.06--0.09 \(e_{\min}\) (6 L or 6 kWh), versus 0.33 for GPT-6 Sol and 0.39 for the full-information reference (Fig.~\ref{fig:energy}b). One trace noted: ``Shortage: 0.63L, which is 10.5\% of the 6L minimum. This is slightly above the 10\% tolerance (0.6L). I should have refueled.\textquotesingle\textquotesingle{} Removing that tolerance clause---with scoring unchanged---cuts Sonnet 5.5\textquotesingle s failures from 6.3 to 2 while leaving non-citing models unchanged. For one family, the tolerance became the target.

\section{SimAct: Simulate Before Acting}\label{sec:framework}

Section 4 suggests that agents rarely fail for lack of information; they misjudge what a plan will do. SimAct tests
this diagnosis by handing consequence prediction to the environment and measuring how much of the gap closes.

\subsection{Design}

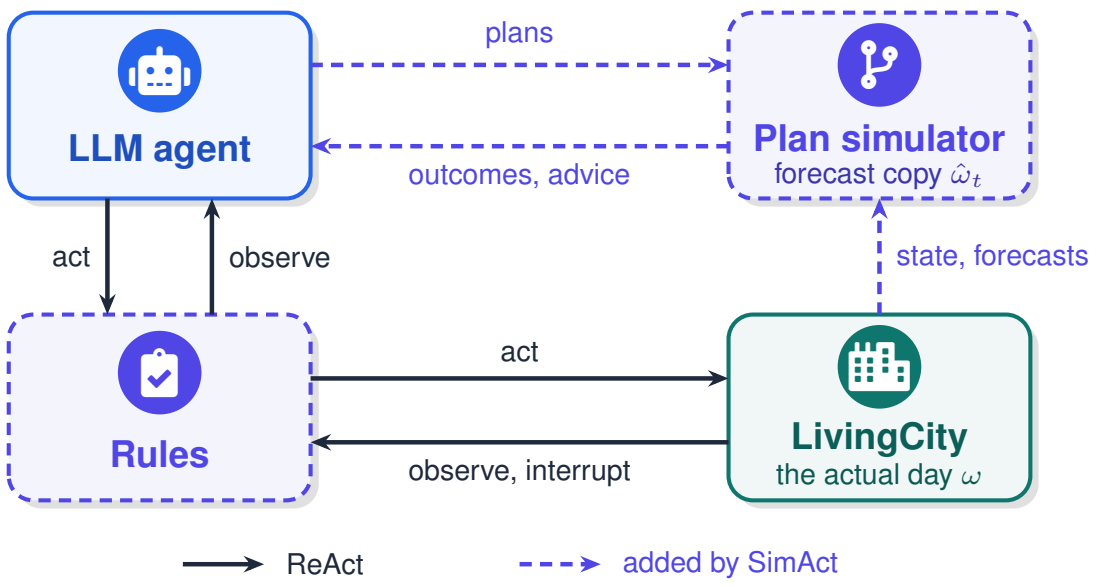
\begin{figure}[t]
\centering
\resizebox{\columnwidth}{!}{\definecolor{ink}{HTML}{1E293B}
\definecolor{frame}{HTML}{64748B}
\definecolor{agentc}{HTML}{2563EB}
\definecolor{succ}{HTML}{4F46E5}
\definecolor{cityc}{HTML}{0F766E}
\begin{tikzpicture}[
  font=\sffamily\footnotesize, text=ink, line width=0.6pt,
  head/.style={font=\sffamily\bfseries\small},
  note/.style={font=\sffamily\scriptsize, align=center},
  arr/.style={-{Stealth[length=5pt, width=4pt]}, line width=1.1pt, draw=ink},
  arrs/.style={-{Stealth[length=5pt, width=4pt]}, line width=1.1pt, draw=succ, dash pattern=on 3pt off 2pt},
  lab/.style={font=\sffamily\scriptsize, align=center},
]
\newcommand{\simbox}[6]{%
  \fill[black!12, rounded corners=6pt, shift={(0.07,-0.07)}] (#1,#2) rectangle (#3,#4);
  \fill[#5!6, rounded corners=6pt] (#1,#2) rectangle (#3,#4);
  \draw[#5, rounded corners=6pt, line width=1pt, #6] (#1,#2) rectangle (#3,#4);}
\newcommand{\badge}[3]{%
  \fill[#2] #1 circle (0.36);
  \node[text=white, font=\large] at #1 {#3};}
\simbox{0}{2.6}{2.6}{4.2}{agentc}{}
\badge{(1.3,3.7)}{agentc}{\faRobot}
\node[head, text=agentc!80!black] at (1.3,3.0) {LLM agent};

\simbox{0}{0}{2.6}{1.6}{succ}{dash pattern=on 3pt off 2pt}
\badge{(1.3,1.1)}{succ}{\faClipboardCheck}
\node[head, text=succ] at (1.3,0.4) {Rules};

\simbox{6.2}{2.6}{8.8}{4.2}{succ}{dash pattern=on 3pt off 2pt}
\badge{(7.5,3.75)}{succ}{\faCodeBranch}
\node[head, text=succ] at (7.5,3.12) {Plan simulator};
\node[note, text=succ!75!black] at (7.5,2.8) {forecast copy $\hat\omega_t$};

\simbox{6.2}{0}{8.8}{1.6}{cityc}{}
\badge{(7.5,1.15)}{cityc}{\faCity}
\node[head, text=cityc!80!black] at (7.5,0.52) {LivingCity};
\node[note, text=cityc!75!black] at (7.5,0.2) {the actual day $\omega$};

\draw[arr] (0.85,2.6) -- node[lab, left] {act} (0.85,1.6);
\draw[arr] (1.75,1.6) -- node[lab, right] {observe} (1.75,2.6);
\draw[arr] (2.6,1.05) -- node[lab, above] {act} (6.2,1.05);
\draw[arr] (6.2,0.5) -- node[lab, below] {observe, interrupt} (2.6,0.5);

\draw[arrs] (2.6,3.75) -- node[lab, above, text=succ] {plans} (6.2,3.75);
\draw[arrs] (6.2,3.05) -- node[lab, below, text=succ] {outcomes, advice} (2.6,3.05);
\draw[arrs] (7.5,1.6) -- node[lab, right, text=succ] {state, forecasts} (7.5,2.6);

\draw[arr] (1.5,-0.55) -- (2.2,-0.55); \node[lab, anchor=west] at (2.25,-0.55) {ReAct};
\draw[arrs] (4.4,-0.55) -- (5.1,-0.55); \node[lab, anchor=west, text=succ] at (5.15,-0.55) {added by SimAct};
\end{tikzpicture}}
\caption{SimAct architecture. Solid: ReAct loop with LivingCity. Dashed: SimAct rolls out candidate plans on a forecast copy (Eq.~\ref{eq:simact}), requiring simulation before acting and blocking early abandonment.}
\label{fig:loops}
\end{figure}

SimAct adds one tool, \texttt{simulate\_plans}, to the ReAct loop (Fig.~\ref{fig:loops}). At time \(t\), the agent proposes up to four candidate plans
\(p_1,\dots,p_k\), each a sequence of actions. LivingCity copies the current state \(s_t\) into a forecast world
\(\hat\omega_t\) holding only what the agent could know: typical traffic, service forecasts, and incidents
already seen. It plays each plan to the end of the day and recommends the lowest-cost plan that completes it:
\begin{equation}\label{eq:simact}
\begin{gathered}
\hat\xi_i = \mathrm{Exec}(p_i; s_t, \hat\omega_t),\\
p^\star = \operatorname*{arg\,min}_{C(\hat\xi_i)=1} G(\hat\xi_i).
\end{gathered}
\end{equation}
The tool also returns each candidate\textquotesingle s lateness, door waiting, spend, and energy margin. The agent may follow
\(p^\star\) or override it. Two rules apply. The agent must simulate before acting at the start and after any deviation. It may not give
up while a simulated plan still completes the day. Rollouts are executed by the environment rather than imagined by the model.
The loop itself is domain-agnostic; only the rollout adapter is domain-specific.

\subsection{Results}

We run all eight agents with SimAct on the same 144 days (Table~\ref{tab:main}, middle block).

\textbf{SimAct helps every agent.} Mean score rises from 34.6 to 48.2. Gains range from 7.9 (DS-V4.1F) to 24.2 (Opus 5.5),
and every bootstrap interval excludes zero. SimAct also re-ranks the leaderboard: Opus 5.5 climbs from
fourth (38.3) to first (62.5), overtaking GPT-6 Sol (59.7). Hard-day scores jump sharply (Opus 5.5: 27.4 to 59.6;
GPT-6 Sol: 41.1 to 55.8), cutting their calm--hard drops to 4.4--5.9 points.

\textbf{It repairs the weaknesses diagnosed in Section 4.} Door waiting drops for every agent: from 26 to 10 minutes for
Sonnet 5.5 and to 4.6 for Opus 5.5. Next-day energy failures fall from 94 to 17 across all eight agents.

\textbf{The repair comes with trade-offs.} Later departures lower on-time appointment rates (GPT-6 Sol: 93\% to 75\%), though
saved waiting outweighs added lateness and hard days improve for every agent. Input tokens rise 1.7--3.3\(\times\).

\subsection{Where the Gain Comes From}

Ablation variants on representative agents isolate each component (Appendix~\ref{app:framework}). Self-prediction without
the simulator drops DS-V4.1F from 29.6 to 24.1 and caps GPT-6 Luna at 34.0. Simulating without recommendation lifts
GPT-6 Luna to 41.2, but DS-V4.1F reaches only 32.2 because it ignores cheaper simulated plans. Enforcing the
recommendation drops GPT-6 Sol from 50.7 to 40.6 because its own departure timing beats the forecast\textquotesingle s. SimAct therefore
only advises. Ranking candidates without score weights---using only completion, lateness, and stated priority---still
captures over half the mean gain. Pricing with \(G\) adds the rest. Thus, the primary bottleneck is predicting plan
outcomes, not proposing plans.

\textbf{Remaining headroom.} Even the best SimAct agent (62.5) trails the full-information reference by 37.5 points. Because
\texttt{simulate\_plans} rolls out on a deterministic forecast copy, it cannot foresee unannounced closures, sudden overflows, or
mid-trip changes. Closing this gap requires uncertainty-aware planning.

\section{Discussion and Conclusion}

Driving agents that plan well on static snapshots do not automatically give a user a good day. DynaDrive-Bench evaluates them in LivingCity, a data-calibrated city that evolves independently, answers back, and reveals surprises only up close. Scoring lived experience above a completion floor exposes gaps that static checks miss. The strongest agent finishes 143 of 144 days yet scores 51 of 100, and eight LLM agents span 15 to 51.

\textbf{What driving agents need.} The primary bottleneck is judging plan consequences rather than sensing live state or reacting to alerts. Agents depart too early, reroute by reflex, and neglect tomorrow\textquotesingle s energy. Rule-based planners show the opposite profile: punctual under calm forecasts but brittle under change. SimAct bridges both by rolling out candidate plans in a forecast copy of the city before acting. It raises all eight agents by 13.6 points on average. Future agents should integrate forward consequence estimation---and uncertainty over sudden disruptions---directly into departure, routing, and energy decisions.

\textbf{What benchmarks need.} Static checks and completed-trip averages misrank methods under disruption. Subtle prompt clauses also matter. Stating a 10\% energy tolerance led one model family to aim at that boundary and leave no margin. This arose from an ordinary task description, not from overt specification gaming \citep{bondarenko_specgaming_2025, thaman_rhb_2026}. Long-horizon benchmarks should pair closed-loop execution with experience-based scoring, multi-run consistency (\(\mathrm{Pass}^k\)) \citep{yao_taubench_2024, kirmayr_carbench_2026}, and tolerance sensitivity checks.

\textbf{Conclusion.} Together, LivingCity, DynaDrive-Bench, and SimAct establish a reproducible testbed and baseline direction for user-centered driving agents in dynamic cities. The simulator, dataset, reference solvers, scoring code, agent loops, and execution replays will be released.

\clearpage  %
\section*{Limitations}
\addcontentsline{toc}{section}{Limitations}

First, LivingCity balances empirical calibration with stress-test design. Congestion curves, route times, and queue profiles are fitted to real traffic and facility data. Remaining parameters, such as stay durations and incident rates, are set by design to sample stressed regimes rather than average days. Travel times are validated against a map service in the ten Chinese cities. In the other six, free-flow speeds follow OpenStreetMap road classes. Across the seven Chinese and five non-Chinese test cities, scores differ by at most 10.2 points, and GPT-6 Sol leads both (55.0 and 44.8).

Second, tasks and user interactions are simplified. Requests come from structured templates, and the simulated user follows deterministic, unambiguous rules rather than open-ended conversation. Extending the benchmark to ambiguous multi-turn dialogue and richer service scenarios, such as unattended charging during activities, is a natural next step.

Third, our evaluation scope has practical compute bounds. All eight agents ran under both ReAct and SimAct on the full test set. Multi-run stability and diagnostic ablations ran on representative models on the 60-day subset.

Fourth, our generalized-cost weights reflect population-level estimates from transport meta-analyses rather than individual user profiles. Varying individual weights or the half-life does not change the leading agent, and extending evaluation to personalized time--money trade-offs remains a promising direction. Map and traffic data are used under their providers\textquotesingle{} terms, which govern which derived parameters we redistribute.

\bibliography{references}

\clearpage  %
\appendix

\raggedbottom

\section{Criteria for Table~\ref{tab:positioning}}\label{app:compare}

We judged each cell from the paper and, where available, the released code; UTP-Bench was judged from its abstract. A
full circle means the property is part of how the benchmark is run and scored. A half circle means it appears in a
weaker form:
\begin{itemize}
\setlength{\itemsep}{1pt}\setlength{\parskip}{0pt}
\item \emph{Closed}: a separate LLM traveler executes the plan (Travel-Sim), or recorded tool calls are replayed (MobilityBench).
\item \emph{Mid-trip}: disruptions, preference changes or passenger requests are scripted in advance (TripTide, CostBench, VehicleArena).
\item \emph{Score}: the robustness of a plan is scored (UTP-Bench), or replayed calls are scored (MobilityBench).
\item \emph{Dynamic}: disruptions are given, uncertainty is sampled, or events block a step (TripTide, UTP-Bench, CostBench).
\item \emph{Partial}: what is hidden is missing information or a private state, not a city revealed on arrival (CAR-bench, VehicleArena, $\tau$-bench, Vending-Bench).
\item \emph{Map}: places are real but there is no road network (TravelPlanner), or data are city-level (TripTide, UTP-Bench).
\item \emph{Energy}: one charging choice (EVGeoQA) or one session (CAR-bench). \emph{Queue}: crowding without queues (UTP-Bench).
\end{itemize}

\section{LivingCity Calibration and Replay}\label{app:sim}\label{app:env}

LivingCity takes its traffic and its car-park and charger queues from real data, and three mechanisms from published
studies (Table~\ref{tab:params}). The remaining values are design choices that set how often a hard situation
arises.

\begin{table*}[!t]
\centering\footnotesize
\caption{Sources of the LivingCity mechanisms.}\label{tab:params}\label{tab:app_calib}
\begin{tabular}{@{}p{0.21\textwidth}p{0.52\textwidth}p{0.21\textwidth}@{}}
\toprule
Mechanism & Value & Source \\
\midrule
\multicolumn{3}{@{}l}{\emph{Fitted to data}} \\
Hourly congestion & Per 2-km cell and road class, fitted to hourly city curves & TomTom (6 cities), Baidu (10) \\
Free-flow speed & Scaled by 0.59--0.95 per Chinese city & AMap, 537 routes \\
Car-park and charger occupancy & Finite-capacity queues; peak quartiles 0.41 / 0.68 / 0.91 (data 0.41 / 0.67 / 0.90) &
Birmingham \citep{stolfi_birmingham_2017}, UrbanEV \citep{li_urbanev_2026} \\
\multicolumn{3}{@{}l}{\emph{From published studies}} \\
Crash duration & 30--90 min on a 500-m segment & \citet{shen_incident_2024} \\
Weather on driving & Rain $\times$1.12, snow $\times$1.30 travel time & \citet{fhwa_weather_2006} \\
Charger faults & 0.20 per charger (0.35 on energy-tight days) & \citet{rempel_dcfc_2022} \\
\multicolumn{3}{@{}l}{\emph{Set by design}} \\
Day-to-day variation & Log-normal multiplier, SD 0.3 per cell and road class & \\
Crashes and closures & 4 crashes on disturbed days, 3 near the planned route; 20\% close the road (40\% on disturbed days) for 20--60 min & \\
Vehicles & Electric: 60\,kWh, 0.16\,kWh/km at 20$^\circ$C. Combustion: 50\,L, 0.085\,L/km. Recommended minimum 6\,kWh or 6\,L & \\
Temperature factor & At $-10$, 0, 10, 20, 30, 35$^\circ$C: electric $\times$1.45, 1.25, 1.10, 1.00, 1.05, 1.15; combustion $\times$1.12, 1.07, 1.03, 1.00, 1.12 (35$^\circ$C); linear in between & \\
Energy per drive & Nominal $\times$ temperature $\times$ a speed factor per 500-m segment; idling also uses energy & \\
Parking & Dearer first hour, 2-h cap at public lots, night rate; reservations; 60\% of turned-away cars spill over within 800\,m & \\
Events, weather, companions & Announced events and rain forecasts on $\sim$25\% of days; companion walk limit 300 or 600\,m, halved in rain & \\
Appointments and messages & Lost after a grace period (e.g.\ 15 min); add, cancel or reschedule an errand, return earlier & \\
\bottomrule
\end{tabular}
\end{table*}

\begin{figure}[!t]
\centering
\includegraphics[width=\linewidth]{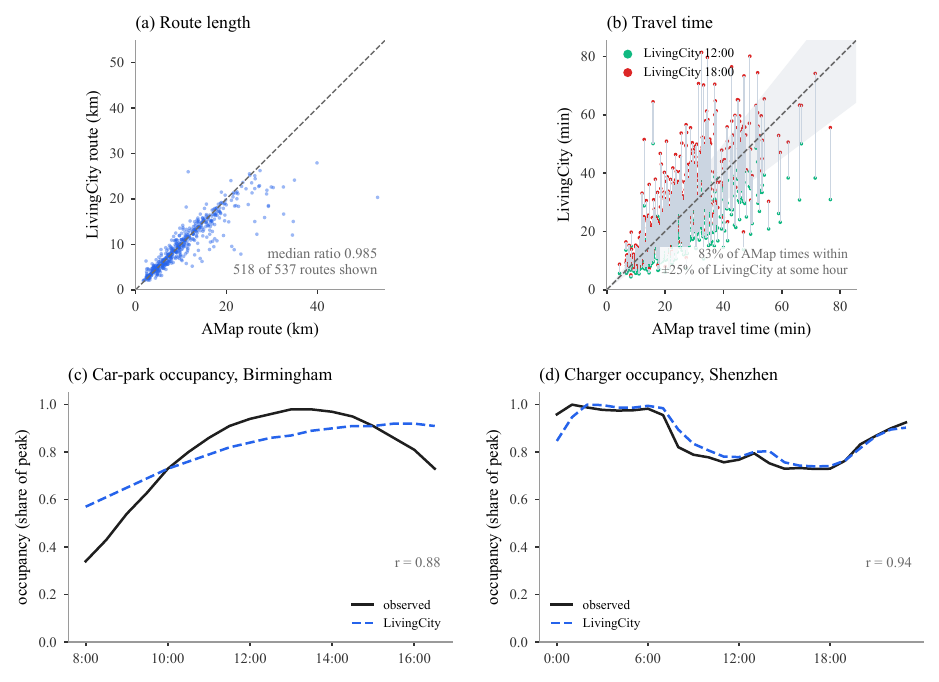}
\caption{LivingCity against real data.}\label{fig:app_validation}
\par\smallskip\parbox{\linewidth}{\footnotesize (a, b) 537 AMap routes in the ten Chinese cities; 19 outliers are
not drawn but are counted. (c, d) Daily occupancy in Birmingham and Shenzhen, observed and simulated.}
\end{figure}

Figure~\ref{fig:app_validation} checks the fitted city against real data. Route lengths and travel times are compared
with 537 AMap routes that were not used for fitting, except for one speed scale per city. The car-park and charger
curves are compared with the data they were fitted to.

\begin{figure*}[!t]
\centering
\includegraphics[width=\textwidth]{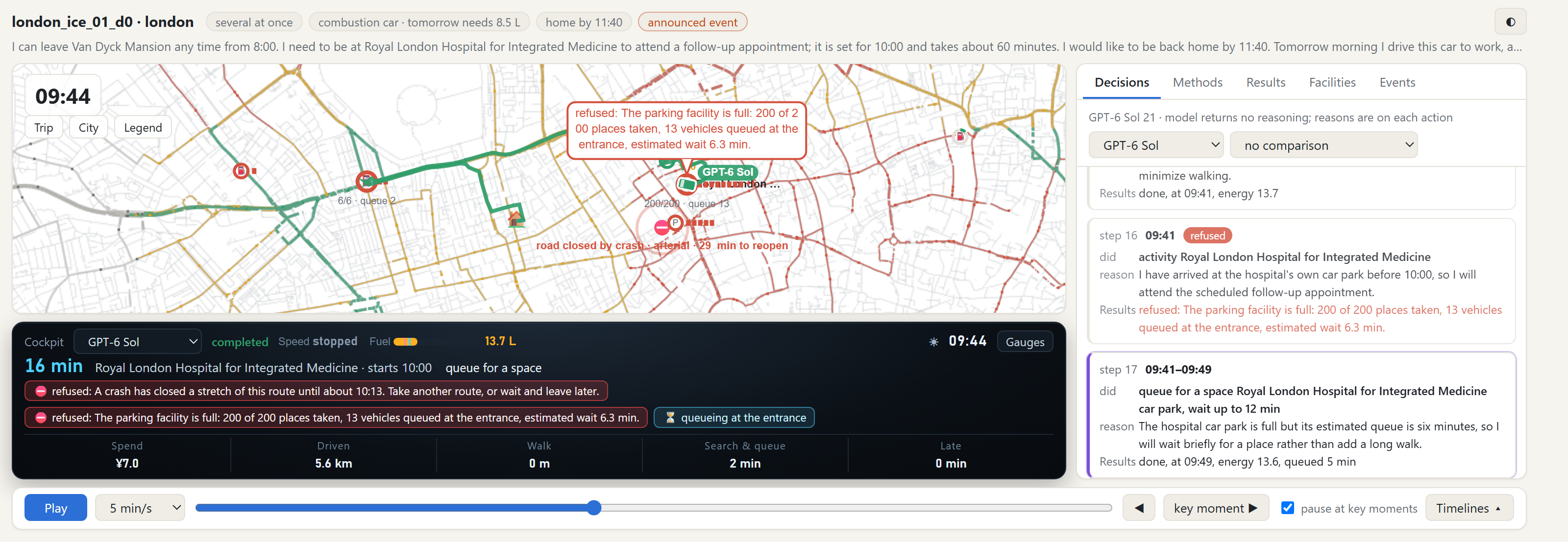}
\caption{One LivingCity day in the replay viewer.}\label{fig:app_city}
\par\smallskip\parbox{\textwidth}{\footnotesize London, 9:44. GPT-6 Sol queues at a full hospital car park after
rerouting around a closed road.}
\end{figure*}

Figure~\ref{fig:app_city} shows one simulated day in the replay viewer. The map colors roads by the day\textquotesingle s
congestion; below it is the cockpit, and on the right are the agent\textquotesingle s steps with its reasons.

\section{Tasks, Test Set and Agent Interface}\label{app:bench}

\subsection{Tasks and the Test Set}\label{app:testset}

The benchmark pairs 192 tasks in 16 cities with sampled days. Table~\ref{tab:app_taxonomy} shows the four task families
with an example request each. The test cities are Beijing, Changchun, Guangzhou, Jinan, Wuhan, Xi'an and Zhengzhou,
with Chinese requests, and London, Los Angeles, New York, Singapore and Sydney, with English requests. Dalian, Nanning,
Shanghai (Pudong) and Toronto are kept for development.

\begin{table*}[!t]
\centering
\footnotesize
\caption{Task families with example requests.}
\label{tab:app_taxonomy}
\setlength{\tabcolsep}{4pt}
\begin{tabular}{p{0.12\textwidth}rp{0.18\textwidth}p{0.32\textwidth}rrrrr}
\toprule
 & & & & \multicolumn{5}{c}{Days by day type} \\
\cmidrule(l){5-9}
Task family & Days & Decisions beyond navigation & Example request (abridged) & Wea. & Tra. & Par. & Ene. & Sev. \\
\midrule
Named errand & 49 & when to leave, which way, where to park, when to refuel or charge & \emph{``I'm free to drive out from Verney House from 9:00 onwards. I need to go to Camden Town Library \ldots\ (about 40 minutes, aiming to arrive by 9:40). I aim to be home before 11:00. I also have to drive to the office tomorrow morning, roughly 25 km \ldots\ I have my two young kids with me.''} & 10 & 9 & 9 & 13 & 8 \\
Need-based errand & 34 & + which place of the requested kind & \emph{``\ldots\ I need to go to a supermarket nearby to do the grocery shopping (about 40 minutes, aiming to arrive by 17:30). I aim to be home before 19:10 \ldots\ roughly 45 km one way.''} & 7 & 8 & 6 & 4 & 9 \\
Fixed-time appointment & 48 & + arriving neither late (lost after the grace period) nor needlessly early & \emph{``\ldots\ I need to be at Royal London Hospital for Integrated Medicine to attend a follow-up appointment; it is set for 10:00 and takes about 60 minutes. I would like to be back home by 11:40 \ldots''} & 8 & 11 & 11 & 10 & 8 \\
Multi-errand day & 13 & + ordering and timing two activities and the energy across both & \emph{``\ldots\ attend a follow-up appointment (set for 18:10, about 60 minutes). I also need to go to a park nearby to take a walk (about 50 minutes, aiming to arrive by 19:50). I aim to be home before 22:00 \ldots''} & 3 & 1 & 4 & 2 & 3 \\
\bottomrule
\end{tabular}
\par\smallskip\parbox{\textwidth}{\footnotesize Last five columns: days by type (weather, traffic, parking, energy, several at once).}
\end{table*}

Day types are drawn with weights 40 (normal), 12 (adverse weather), 12 (disturbed traffic), 12 (tight parking), 10
(tight energy) and 14 (several at once), giving 1,496 task--day instances. For each test task we take one stressed day,
choosing only from what is known before any run. A day is kept only if a reference solver completes it (143 by the full-information search and 1 by the fixed plan). The test set and the 60-day stability subset were frozen
with a fixed seed before any model was run.

\subsection{What the Agent Sees and Does}\label{app:obs}\label{app:task}

The agent acts through eight tools (Table~\ref{tab:app_tools}), described in full in the system prompt
(Appendix~\ref{app:prompt}). Seven query the city or end the day and take no simulated time. The eighth,
\texttt{execute\_action}, performs one physical action with a one-sentence reason and moves the clock.

\begin{table}[!t]
\centering\footnotesize
\caption{Tools and physical actions available to the agent.}\label{tab:app_tools}\label{tab:app_actions}
\setlength{\tabcolsep}{3pt}
\begin{tabular}{@{}l>{\raggedright\arraybackslash}p{\linewidth-\widthof{\texttt{search\_locations}}-2\tabcolsep}@{}}
\toprule
Tool / action & Description and return payload \\
\midrule
\multicolumn{2}{@{}l}{\emph{Queries and finishing: take no simulated time}} \\
\texttt{get\_task} & User request text, vehicle specification, required activities, notices, and constraints \\
\texttt{observe\_state} & Current time, location, energy, action history, and all user messages received so far \\
\texttt{search\_locations} & Candidate venues and coordinates for the trip \\
\texttt{search\_services} & Car parks, chargers, and fuel stations with prices, capacities, and hourly forecasts \\
\texttt{query\_routes} & Up to three drive (or walk) routes per pair, free-flow/forecast times, energy, tolls, closures \\
\texttt{query\_realtime} & Live observations: facility at hand, telemetry-enabled facilities, live drive times \\
\texttt{finish\_trip} & Terminates the episode as \emph{completed}, \emph{no feasible plan}, or \emph{abort} \\
\midrule
\multicolumn{2}{@{}l}{\emph{Actions via \texttt{execute\_action}: advance the clock}} \\
Moving & \texttt{drive} or \texttt{walk} a route; \texttt{continue} or reroute a paused drive from a junction; \texttt{board} \\
Parking & \texttt{park}, \texttt{unpark}; \texttt{queue\_for\_space} at a full lot (with max wait); \texttt{reserve} a space/charger \\
Energy & \texttt{charge} or \texttt{refuel} a target amount (with optional max wait); \texttt{stop\_charge}; \texttt{continue} \\
Errands & \texttt{activity} at a venue; \texttt{wait}; \texttt{contact} an appointment host to request a grace extension \\
Companion & \texttt{drop\_off} and \texttt{pick\_up} an elderly parent or child at the venue entrance \\
\bottomrule
\end{tabular}
\end{table}

The agent can always read forecasts, but sees the realized city only as a driver would:
\begin{itemize}
\setlength{\itemsep}{1pt}\setlength{\parskip}{0pt}
\item \textbf{Always available:} typical hourly congestion and occupancy, weather forecasts, announced events and road
closures, routes, distances and tariffs.
\item \textbf{Live, by remote query:} drive times including reported crashes; occupancy of about 20\% of car parks (5--15
min late, with noise) and 80\% of charging stations.
\item \textbf{Only on arrival or approach:} unannounced crashes within about 1.5\,km, and whether a car park without live
data is full.
\item \textbf{Never:} the realized day itself, including future crashes, future arrivals and upcoming user messages.
\end{itemize}
Besides the drive interruptions listed in the main text, charging pauses on a tariff change, a charger fault or a user
message. Running out of energy on the road ends the day. Figure~\ref{fig:trace} shows one exchange with two
interruptions: a crash ahead, then a user message on arrival.

\begin{figure}[!t]
\begin{tcolorbox}[enhanced, colback=black!3, colframe=black!45, boxrule=0.5pt, arc=2pt, left=4pt, right=4pt, top=3pt,
  bottom=3pt, fontupper=\footnotesize\raggedright]
\textbf{16:03} \texttt{execute\_action(refuel, S2, 5\,L)}\\
\emph{reason:} CNY 7.69 per liter and a short queue; the minimum 5\,L covers today and the reserve for tomorrow\textquotesingle s
45\,km.\par\smallskip
\textbf{16:09} \texttt{execute\_action(drive, S2\,$\to$\,venue car park)}\\
\emph{reason:} fuel is enough and the car park may fill before the match; take the fastest route now.\par\smallskip
\textcolor{red!70!black}{$\leftarrow$ \textbf{16:16 interrupted at junction J1}: \texttt{incident\_ahead}: \textquotedblleft an incident
blocks the road ahead (about 1\,km); traffic there is at a standstill\textquotedblright; continue, or drive another route
from J1.}\par\smallskip
\textbf{16:16} \texttt{query\_routes(origin=J1)} $\to$ fastest 16.5\,min, no expressway 19.1, alternative 19.3 (forecast)\par\smallskip
\textbf{16:16} \texttt{execute\_action(drive, J1\,$\to$\,venue, no expressway)}\\
\emph{reason:} the main road ahead is stalled by the incident; go around it without expressways, to arrive before
17:30.\par\smallskip
\textcolor{red!70!black}{$\leftarrow$ \textbf{16:41 arrived}; the user says: \textquotedblleft Something came up at home; I
need to be back by 18:15.\textquotedblright}
\end{tcolorbox}
\caption{One exchange from a test run (GPT-6 Sol, abridged).}
\label{fig:trace}
\end{figure}

\section{Scoring Justification and Sensitivity}\label{app:scoring}

Changing the half-life or any single weight does not change the leading agent (Table~\ref{tab:app_sensitivity}).
Table~\ref{tab:app_weights} summarizes the generalized-cost weights (Eq.~\ref{eq:gcost}) alongside transport meta-analysis
evidence \citep{wardman_multipliers_2026, wardman_vtts_2023}.

\begin{table}[!t]
\centering
\footnotesize
\setlength{\tabcolsep}{2pt}
\caption{Score under other half-lives and weights.}\label{tab:app_sensitivity}
\fitwidth{%
\begin{tabular}{lrrrrrrrrrr}
\toprule
Variant & Sol & GLM & Terra & Opus & Sonnet & DS & Luna & MiMo & Fixed & Online \\
\midrule
Base & \textbf{50.7} & 40.2 & 39.3 & 38.3 & 36.4 & 29.6 & 27.5 & 15.1 & 36.5 & 41.1 \\
$H=5\%$ & \textbf{40.9} & 32.7 & 31.2 & 32.8 & 31.0 & 23.5 & 24.4 & 10.3 & 28.6 & 31.9 \\
$H=20\%$ & \textbf{62.6} & 49.8 & 49.5 & 45.7 & 43.7 & 38.0 & 31.5 & 22.6 & 46.1 & 52.3 \\
$H=40\%$ & \textbf{74.3} & 61.1 & 61.4 & 55.5 & 53.1 & 48.8 & 36.1 & 33.1 & 55.9 & 63.4 \\
5 min & \textbf{40.5} & 32.0 & 30.5 & 33.3 & 31.1 & 23.5 & 24.9 & 10.5 & 29.5 & 32.5 \\
Wait 1.5 & \textbf{50.7} & 40.5 & 38.9 & 38.0 & 35.7 & 29.7 & 27.6 & 15.2 & 36.5 & 41.1 \\
Wait 2.5 & \textbf{51.2} & 40.1 & 39.8 & 38.5 & 37.2 & 29.6 & 27.5 & 15.1 & 36.6 & 41.3 \\
Door 0.6 & \textbf{55.7} & 42.5 & 43.5 & 42.7 & 42.0 & 32.8 & 30.6 & 16.9 & 37.1 & 41.7 \\
Door 2 & \textbf{49.5} & 39.9 & 38.4 & 37.5 & 35.4 & 28.9 & 27.0 & 14.7 & 36.1 & 40.7 \\
Late 2 & \textbf{50.7} & 40.5 & 39.6 & 38.4 & 36.4 & 29.5 & 27.5 & 15.2 & 36.9 & 41.5 \\
Late 3 & \textbf{50.9} & 39.9 & 39.0 & 38.2 & 36.4 & 29.7 & 27.5 & 14.9 & 36.0 & 40.6 \\
VOT $\times$2 & \textbf{49.3} & 38.7 & 38.0 & 36.6 & 35.8 & 28.8 & 26.3 & 14.5 & 36.4 & 40.8 \\
VOT $\times$0.5 & \textbf{53.2} & 43.0 & 40.9 & 40.2 & 37.3 & 31.2 & 28.7 & 16.6 & 37.6 & 42.2 \\
\bottomrule
\end{tabular}}
\par\smallskip\parbox{\linewidth}{\footnotesize Each row changes one setting of the base score; bold marks the best
agent. $H$: half-life. \emph{5 min}: the score halves per 5 extra minutes. \emph{Wait}: weight of walking and waiting.
\emph{Door}: weight of door waiting. \emph{VOT}: values of time. All runs use ReAct.}
\end{table}

\begin{table}[!t]
\centering
\footnotesize
\setlength{\tabcolsep}{4pt}
\caption{Weights of the generalized cost.}\label{tab:app_weights}
\fitwidth{\begin{tabular}{@{}lrl@{}}
\toprule
Cost to the user & Weight & Evidence \\
\midrule
Driving (min) & 1 & reference unit \\
Walking (min) & 2 & 1.67 \\
Searching or queuing (min) & 2 & 1.60 \\
Waiting for energy, other waits (min) & 2 & 1.72 \\
Door wait before a fixed start (min) & 1.5 & early 0.96; wait 1.72 \\
Lateness (min) & 2.4 & 2.23; urban car 2.6--3.0 \\
Value of time, Chinese cities & CNY 50/h & GDP per head \\
Value of time, other cities & CNY 128--176/h & GDP per head \\
\bottomrule
\end{tabular}}
\par\smallskip\parbox{\linewidth}{\footnotesize Evidence gives meta-analysis means \citep{wardman_multipliers_2026}.
Values of time follow the income elasticity of 0.696 in \citet{wardman_vtts_2023}, at the level for personal trips.}
\end{table}

\paragraph{Same-day references.} All three references search the same family of plans. A plan fixes the order of the
activities, a place for each open request, a car park for each activity and the energy stops.
\begin{itemize}
\setlength{\itemsep}{1pt}\setlength{\parskip}{0pt}
\item The \emph{fixed plan} searches the forecast, with departures every 15 minutes up to 120 minutes after the earliest
time. It ranks plans by completion, lateness, the user\textquotesingle s priority and walking. It runs without live
queries, queues at full car parks, and replans on the forecast only when the user changes the day.
\item The \emph{online replanner} replans the same plan with live conditions at every interruption and user message.
\item The \emph{full-information reference} \(\xi^*\) searches the realized day, knowing every crash, arrival and
message, with departures every 10 minutes. It ranks plans by completion, then by generalized cost (Eq.~\ref{eq:gcost}).
On one of the 144 days it finds no plan and follows the fixed plan, which completes that day.
\end{itemize}
Because the search is restricted, agents sometimes beat the reference: GPT-6 Sol does on 20 of its 143 completed days,
and such a day scores 100.

\paragraph{Cost and reserve.} Minutes are counted from leaving home to arriving home, by what the person is doing
(Table~\ref{tab:app_weights}). Time on the activities and waiting at home before leaving are not counted. Net spend
values the energy drawn from the tank at the median price on the task\textquotesingle s map, so energy bought at the
usual price is neutral. Spend in the five non-Chinese test cities is converted to yuan at fixed rates. Tomorrow\textquotesingle s
distance is 25, 35 or 45\,km. The reserve (Eq.~\ref{eq:reserve}) uses the consumption and temperature factors in
Table~\ref{tab:params}. It is rounded up to the next half unit, and energy added after reaching home does not count.

\section{Experimental Details and Additional Diagnostics}\label{app:results}\label{app:models}

\begin{table*}[!t]
\centering
\footnotesize
\setlength{\tabcolsep}{3.5pt}
\captionof{table}{Score (0--100) by condition and task structure.}\label{tab:app_conditions}
\fitwidth{%
\begin{tabular}{lrrrrrrrrrrrrrrrr}
\toprule
~ & \multicolumn{5}{c}{Day type} & \multicolumn{4}{c}{Task family} & ~ & ~ & ~ & \multicolumn{4}{c}{Starting energy} \\
\cmidrule(lr){2-6}\cmidrule(lr){7-10}\cmidrule(lr){14-17}
Method & \shortstack{Parking\\{}tight (30)} & \shortstack{Several\\{}(28)} & \shortstack{Weather\\{}(28)} & \shortstack{Energy\\{}tight (29)} & \shortstack{Traffic\\{}(29)} & \shortstack{Named\\{}errand (49)} & \shortstack{Need-\\{}based (34)} & \shortstack{Appoint-\\{}ment (48)} & \shortstack{Multi-\\{}errand (13)} & \shortstack{Mid-trip\\{}message (51)} & \shortstack{Hard\\{}days (50)} & \shortstack{Calm\\{}days (94)} & \shortstack{Ample\\{}(31)} & \shortstack{Tight\\{}(57)} & \shortstack{Low\\{}(39)} & \shortstack{Very low\\{}(17)} \\
\midrule
GPT-6 Sol & 57 & 51 & 45 & 54 & 47 & 59 & 67 & 31 & 49 & 45 & 41 & 56 & 66 & 47 & 48 & 43 \\
GLM-5.3F & 39 & 43 & 32 & 51 & 37 & 55 & 56 & 17 & 29 & 36 & 32 & 44 & 55 & 38 & 36 & 30 \\
GPT-5.6 Terra & 43 & 41 & 38 & 41 & 34 & 52 & 59 & 16 & 28 & 30 & 33 & 43 & 55 & 41 & 33 & 22 \\
Opus 5.5 & 42 & 26 & 33 & 45 & 44 & 59 & 61 & 3 & 31 & 34 & 27 & 44 & 49 & 39 & 32 & 29 \\
Sonnet 5.5 & 40 & 31 & 36 & 38 & 37 & 56 & 60 & 3 & 22 & 30 & 24 & 43 & 50 & 37 & 29 & 29 \\
DS-V4.1F & 26 & 26 & 35 & 38 & 23 & 35 & 51 & 12 & 20 & 24 & 23 & 33 & 48 & 29 & 20 & 20 \\
GPT-6 Luna & 28 & 27 & 36 & 21 & 26 & 39 & 52 & 3 & 8 & 19 & 23 & 30 & 53 & 34 & 10 & 1 \\
MiMo-V2.6F & 14 & 20 & 19 & 18 & 5 & 25 & 16 & 7 & 7 & 19 & 13 & 16 & 18 & 15 & 14 & 14 \\
Fixed plan & 46 & 24 & 33 & 46 & 32 & 45 & 41 & 25 & 34 & 34 & 20 & 45 & 39 & 44 & 24 & 36 \\
Online replanner & 48 & 30 & 34 & 55 & 37 & 48 & 42 & 37 & 26 & 37 & 24 & 50 & 48 & 41 & 32 & 48 \\
\bottomrule
\end{tabular}}
\par\smallskip\parbox{\linewidth}{\footnotesize Number of days in parentheses. Days by how many of the eight agents completed them (3: 1, 4: 6, 5: 9, 6: 29, 7: 43, 8: 56); no day was completed by fewer than four agents.}
\end{table*}

\begin{table*}[!t]
\centering
\footnotesize
\setlength{\tabcolsep}{3.5pt}
\captionof{table}{Why days were not completed, and the days lost only on tomorrow's energy.}\label{tab:app_failures}
\fitwidth{%
\begin{tabular}{lrrrrrrrrrrrrr}
\toprule
~ & \multicolumn{10}{c}{Reason (first that applies, in column order)} & \multicolumn{3}{c}{Lost only on next-day energy} \\
\cmidrule(lr){2-11}\cmidrule(lr){12-14}
Method & \shortstack{Incom-\\{}plete} & \shortstack{Out of\\{}energy} & \shortstack{Appt.\\{}lost} & Closed & \shortstack{Gave\\{}up} & \shortstack{Tool\\{}budget} & \shortstack{Not\\{}home} & \shortstack{Errands\\{}left} & \shortstack{Request\\{}ignored} & \shortstack{Next-day\\{}energy} & Days & \shortstack{Mentioned\\{}tomorrow} & \shortstack{Stopped\\{}for energy} \\
\midrule
GPT-6 Sol & 1 &  &  &  &  &  &  & 1 &  &  & -- & -- & -- \\
GLM-5.3F & 6 &  &  &  &  &  & 1 & 3 &  & 2 & 2 & 100\% & 50\% \\
GPT-5.6 Terra & 6 &  &  &  & 1 &  &  &  &  & 5 & 5 & 100\% & 60\% \\
Opus 5.5 & 10 &  &  &  &  &  &  & 2 &  & 8 & 8 & 100\% & 25\% \\
Sonnet 5.5 & 13 & 1 &  &  &  &  &  & 2 &  & 10 & 10 & 100\% & 0\% \\
DS-V4.1F & 19 & 1 &  &  & 1 &  &  & 5 &  & 12 & 12 & 67\% & 17\% \\
GPT-6 Luna & 71 &  &  &  & 23 &  &  & 6 & 1 & 41 & 41 & 56\% & 10\% \\
MiMo-V2.6F & 29 & 1 &  &  &  & 1 & 1 & 8 & 2 & 16 & 16 & 25\% & 19\% \\
Fixed plan & 31 &  & 21 &  &  &  &  & 3 & 1 & 6 & -- & -- & -- \\
Online replanner & 19 &  & 16 &  &  &  &  & 3 &  &  & -- & -- & -- \\
\bottomrule
\end{tabular}}
\par\smallskip\parbox{\linewidth}{\footnotesize Closed: venue or car park closed. Mentioned tomorrow: the agent's text refers to the next day (keyword match). Blank cells are zero.}
\end{table*}

\paragraph{Runs.} The budget per day is given in the main text. A run also stops after a 300-second request timeout,
and each provider keeps its default sampling. Models are reached through commercial gateways, with prompt caching where
supported. A gateway can add instructions of its own, so we compared the billed input tokens of every first request with
what was sent; none showed added instructions. The Claude models have about 21\% of their actions refused, mostly for an
extra \texttt{location\_id} or an invented action name; they correct these within a few attempts.

\paragraph{Results by condition and failure causes.}\label{app:conditions}\label{app:failures} Days with several stresses
at once and hard days separate methods most (Table~\ref{tab:app_conditions}). Fixed-time appointments and multi-errand
days are hardest for every agent, especially Claude models whose early door waits drop appointment scores to 3, whereas
rule-based references drop mainly by losing appointments (Table~\ref{tab:app_failures}). Across next-day-only failures,
GPT-5.6 Terra, GLM-5.3F, and both Claude models explicitly mention tomorrow\textquotesingle s commute in 100\% of traces.

\begin{figure}[!t]
\centering
\includegraphics[width=\linewidth]{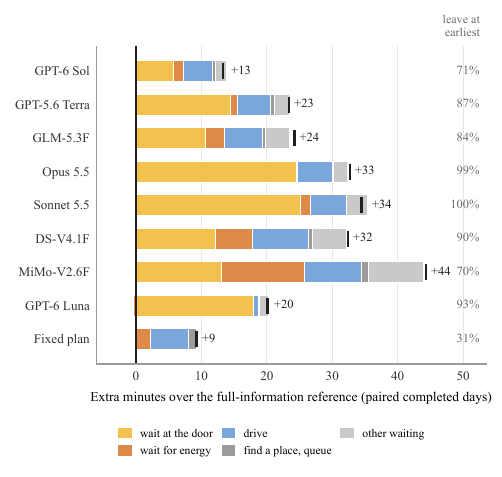}
\caption{Extra minutes beyond the same-day full-information reference on paired completed days, with earliest-departure share on the right.}\label{fig:timecost}
\end{figure}

\paragraph{Where the time goes and stability.}\label{app:adaptation} Figure~\ref{fig:timecost} splits each method\textquotesingle s
extra minutes on completed days by user state: Claude models lose time at the door, whereas DS-V4.1F and MiMo-V2.6F
accumulate driving, energy-waiting, and other delays. On the 60-day subset, three runs of each of four agents differ by at
most 6 points: GPT-6 Sol 49--52, Sonnet 5.5 34--38, GPT-6 Luna 26--31 and MiMo-V2.6F 15--18. Completing all three runs is
rarer: Pass$^3$ is 96.7\%, 73.3\%, 43.3\% and 56.7\%. Without the sentence that allows a 10\% energy shortfall,
MiMo-V2.6F and GPT-6 Luna stay within or near their baseline ranges.

\begin{figure}[!t]
\centering
\includegraphics[width=\linewidth]{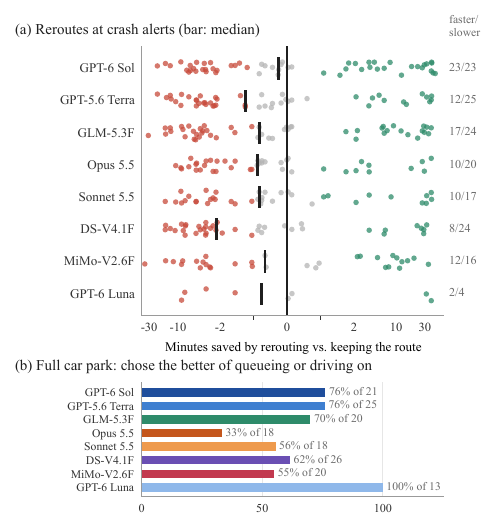}
\caption{Same-day forks at crash alerts and full car parks.}\label{fig:adapt}
\end{figure}

\paragraph{Same-day forks.}\label{app:alerts} Because LivingCity is deterministic, any recorded
day can be forked at a decision and both branches replayed on the identical day. At a crash alert, one branch keeps the
route and the other takes the agent\textquotesingle s detour; they are compared at that drive\textquotesingle s
destination. At a full car park, the branches queue or drive to each other car park within walking distance, and are
compared at the door. Differences under one minute count as ties (Figure~\ref{fig:adapt}).

\section{SimAct Variants and Prompt Addendum}\label{app:framework}

Table~\ref{tab:app_framework} lists every variant of the agent loop we ran, each once on the 144 test days. These runs allow 4,096 output tokens per turn instead of 2,048, since one turn may write several plans. Other budgets match the ReAct runs.
\emph{Self-prediction} has no simulator: the agent writes its own forecast of each candidate, and the harness returns
only predictions that contradict themselves. \emph{Simulator, no recommendation} returns the rollouts but names no plan.
\emph{Recommendation enforced} refuses actions outside the recommended plan. Variants \emph{without score weights} rank
candidates by completion, lateness, then the user\textquotesingle s priority, then the other priority, then walking.

\begin{table*}[!t]
\centering\footnotesize
\setlength{\tabcolsep}{4pt}
\caption{SimAct variants: score (days completed).}
\label{tab:app_framework}
\fitwidth{\begin{tabular}{@{}lrrrrrrrr@{}}
\toprule
Variant & Sol & GLM & Terra & Opus & Sonnet & DS & Luna & MiMo \\
\midrule
ReAct & 50.7 (143) & 40.2 (138) & 39.3 (138) & 38.3 (134) & 36.4 (131) & 29.6 (125) & 27.5 (73) & 15.1 (115) \\
Self-prediction, no simulator & -- & -- & -- & -- & -- & 24.1 (134) & 34.0 (91) & -- \\
Simulator, no recommendation & -- & -- & -- & -- & -- & 32.2 (140) & 41.2 (116) & -- \\
Recommendation enforced & 40.6 (139) & 51.1 (138) & -- & -- & 44.7 (130) & 39.4 (137) & 39.3 (120) & -- \\
Enforced, no score weights & -- & -- & -- & -- & -- & 34.3 (139) & 32.7 (112) & -- \\
\textbf{SimAct} & 59.7 (140) & 53.5 (138) & 53.1 (140) & 62.5 (142) & 44.6 (135) & 37.5 (138) & 44.3 (115) & 30.9 (134) \\
Advice, no score weights & -- & 47.9 (141) & -- & -- & -- & 30.3 (136) & 39.8 (139) & -- \\
\bottomrule
\end{tabular}}
\par\smallskip\parbox{\textwidth}{\footnotesize --: not run.}
\end{table*}

\texttt{simulate\_plans} takes one to four candidates, each a list of actions in the format of \texttt{execute\_action};
steps the environment fills in, such as walking back to the car, may be left out. For each candidate it returns
completion, lateness, door waiting, trip time, spend, walking, energy at home minus tomorrow\textquotesingle s need, a
list of problems and the forecast generalized cost, and it names the recommended candidate. An action without a fresh
simulation is refused at the start and after a deviation; after three refusals the action goes through. Giving up is
refused, at most three times per run, while the latest simulation holds a candidate that completes the day. The task
description is unchanged; SimAct appends the following workflow prompt to the system prompt in Appendix~\ref{app:prompt}:

\begin{tcolorbox}[enhanced, breakable, colback=black!3, colframe=black!40, boxrule=0.4pt, arc=2pt, left=5pt, right=5pt,
  top=4pt, bottom=4pt, fontupper=\footnotesize\raggedright]
\textbf{WORKFLOW: look ahead, then act.} You have a tool \texttt{simulate\_plans}.\par\smallskip
1. Before the first action, read the task. Write 2--4 complete candidate plans that reach the end of the task. Make them
differ where it matters for this task: timing, order, which place or option, which resources to use and when, and the
route. At least two candidates should start at clearly different times.\par\smallskip
2. Simulate them with \texttt{simulate\_plans}. It plays each plan from the current state on the public forecast and
reports whether it completes the task, whether it meets the task\textquotesingle s requirements, the problems it runs
into (lateness, waiting and other costs included), and the predicted outcome. It executes nothing, and the actual
situation may differ from the forecast.\par\smallskip
3. Choose. Prefer plans that meet the task\textquotesingle s requirements. Keep a margin only where a requirement could
fail if things turn out worse than forecast. Arriving early and waiting is a cost to the user, as is any other wait or
extra spend. Among the plans that meet the requirements, compare problems and costs by the priorities the task states.
If no plan meets the requirements, change the plans where the problems point and simulate again. A plan that completes
the task with problems still serves the user far better than abandoning the task. Declare the task infeasible only when
plans that differ in these ways all fail to complete it.\par\smallskip
4. Execute the chosen plan one action at a time.\par\smallskip
5. Simulate again from the current state when the situation changes: an action is interrupted or rejected, new
instructions arrive, or what you observe drifts from your simulation. Then choose by the same rules.\par\smallskip
The harness asks for a simulation before the first action, after each interruption, rejection or new instruction, and
before giving up. It refuses to give up while your latest simulation from the current state shows a plan that completes
the task.\par\smallskip
In this workflow the harness advises in step 3. Each simulated plan comes with \texttt{plan\_cost}, its overall cost to
the user on the forecast. Among all plans you simulated from the current state that complete the task and meet its
requirements, the harness names the one with the lowest \texttt{plan\_cost} as \texttt{framework\_selection}. The
decision stays yours: the forecast can miss what you know or observe, and no action is refused for leaving that plan.
\end{tcolorbox}

\section{Data, Licenses and Intended Use}\label{app:data}\label{app:release}\label{app:sources}

Every external source enters the benchmark only through the city maps or fitted simulator parameters
(Table~\ref{tab:app_sources}).

\begin{table*}[!t]
\centering\footnotesize
\caption{Data sources and their terms.}\label{tab:app_sources}
\begin{tabular}{@{}>{\raggedright\arraybackslash}p{0.3\textwidth}>{\raggedright\arraybackslash}p{0.48\textwidth}>{\raggedright\arraybackslash}p{0.16\textwidth}@{}}
\toprule
Source & Used for & License or terms \\
\midrule
OpenStreetMap (\url{openstreetmap.org}) & Road networks of all 16 cities; venues and facilities in the six non-Chinese cities & ODbL 1.0 \\
AMap (\url{lbs.amap.com}) & Venues and facilities in the ten Chinese cities; 537 routes for calibration and validation & AMap platform terms \\
TomTom Traffic Index (\url{tomtom.com/traffic-index}) & Hourly congestion in six cities & Website terms \\
Baidu Maps traffic index (\url{jiaotong.baidu.com}) & Hourly congestion in ten cities & Website terms \\
Parking Birmingham \citep{stolfi_birmingham_2017} & Car-park occupancy, 30 car parks, 2016 & CC BY 4.0 \\
UrbanEV \citep{li_urbanev_2026} & Charger occupancy, duration and prices, Shenzhen, 2022--2023 & CC0 1.0 \\
\bottomrule
\end{tabular}
\end{table*}

We will release the LivingCity simulator, all 1,496 task--day instances with the test set and the stability subset,
the three reference solvers, the scoring code, the agent harness with its task description, the recorded runs of all
eight agents, and the replay viewer. Road graphs derived from OpenStreetMap will be released under ODbL. Raw TomTom,
Baidu and AMap data will not be redistributed; we will release the scripts that fetch and fit them, with the fitted
parameters.

The benchmark is meant for evaluating language-model agents in simulation. LivingCity follows typical patterns, not
traffic on any real day, and its outputs should not guide real driving. The experiments involve no human participants.
Homes and venues are public places from map data, requests come from templates, and no task describes a real person.
The recorded runs contain model outputs obtained through commercial APIs under each provider\textquotesingle s terms.

\FloatBarrier
\section{Task Description}\label{app:prompt}

Every agent received the following system prompt, reproduced in full. Its section \emph{How the day is judged} calls a
day that \textquotesingle succeeds' what the paper calls completed; the paper\textquotesingle s score also prices the
time and money the day cost.

\begin{tcolorbox}[breakable, enhanced, colback=black!3, colframe=black!45, boxrule=0.5pt, arc=2pt,
  left=5pt, right=5pt, top=4pt, bottom=4pt, title={System prompt given to every agent (complete)},
  fonttitle=\small\bfseries, coltitle=black, colbacktitle=black!10, fontupper=\footnotesize\raggedright]
\par\smallskip You are a driving agent. You take one person's trip through a simulated city on their behalf: when to leave, which routes to drive, where to park, where and how much to charge or refuel, and how to adapt while the day unfolds. The city changes over time and answers what you do; you see it only as a driver would.\par
\par\smallskip\textbf{How a run works}\par
\begin{itemize}[leftmargin=1.1em,itemsep=1pt,topsep=1pt,parsep=0pt]
\item Tools: \texttt{get\_task} (the user's request in their words, the vehicle, the activities), \texttt{observe\_state}, \texttt{search\_locations}, \texttt{search\_services}, \texttt{query\_routes}, \texttt{query\_realtime}, \texttt{execute\_action} and \texttt{finish\_trip}. Query tools take no time; \texttt{execute\_action} moves the clock.
\item Send at most one \texttt{execute\_action} per turn; it carries reason, one sentence on what you observed and why you act now. After it, read the returned state before deciding again.
\item A rejected action changes nothing and says why; correct it and go on. A physical failure (running out of energy, a car locked in a closed car park) ends the day.
\item End with \texttt{finish\_trip}: completed, \texttt{no\_feasible\_plan} or abort. The day is judged by what actually happened, not by what you declare.
\item Units: time in minutes since midnight on the task clock, distance in metres, energy in kWh (electric) or litres (combustion), prices in \texttt{conditions.currency}.
\end{itemize}
\par\smallskip\textbf{Moving around}\par
\begin{itemize}[leftmargin=1.1em,itemsep=1pt,topsep=1pt,parsep=0pt]
\item Only listed route ids can be driven or walked; a pair of places has up to three drive routes (label: fastest, \texttt{major\_roads}, \texttt{no\_expressway}, alternative). \texttt{duration\_min} is the free-flow time; \texttt{query\_routes} gives \texttt{forecast\_duration\_min} for a departure time under typical congestion. Congestion differs by district, road type, time of day and day.
\item Every route shown carries closed (true or false). A closed route passes announced roadworks (\texttt{closed\_hours}); driving it is rejected.
\item \texttt{get\_task} may list restrictions: rules that close a zone to some vehicles at some hours (a drive through it is rejected), charge a daily fee for driving in a zone (paid once), or close a stretch of road for roadworks.
\item Some routes have tolls. Weather (\texttt{conditions.weather}) slows driving and walking.
\end{itemize}
\par\smallskip\textbf{What you can know}\par
\begin{itemize}[leftmargin=1.1em,itemsep=1pt,topsep=1pt,parsep=0pt]
\item Forecasts (the usual pattern by time of day) are always available: congestion through \texttt{query\_routes}, and for each facility \texttt{forecast\_by\_time} (occupancy, \texttt{full\_prob}, queue, expected waits).
\item \texttt{query\_realtime} shows what can be seen now: the facility where the car or the person is; facilities that publish live data (\texttt{capacity\_model.live\_data}; the reading is \texttt{live\_delay\_min} minutes old and roughly right); and drive times from the vehicle's location if leaving now. Live drive times include only incidents you already know of, and are estimates that can be somewhat off.
\item Road incidents are not forecast. You learn of one when your drive comes close to it (the drive pauses) or after driving through it.
\item A car park without live data shows whether it is full only when you are at its entrance.
\item Today is not exactly the usual day: activities can take longer or shorter than \texttt{duration\_min} (\texttt{duration\_spread} is the typical spread), and the user may say something during the trip.
\end{itemize}
\par\smallskip\textbf{Energy}\par
\begin{itemize}[leftmargin=1.1em,itemsep=1pt,topsep=1pt,parsep=0pt]
\item Energy used on a drive = \texttt{nominal\_energy\_20c} x the temperature factor for today x \texttt{speed\_energy\_factor}. Waiting in the car uses some energy too, more in cold or heat.
\item The vehicle lists \texttt{recommended\_minimum\_energy}. The user also says how far the car will be driven tomorrow morning. The energy needed on arriving home is that distance x \texttt{nominal\_consumption\_per\_km\_20c} x the temperature factor for today, plus \texttt{recommended\_minimum\_energy} (rounded up to the next 0.5).
\item Charging (electric) and refuelling (combustion) happen at energy services; amount is the energy bought. Charging power follows the vehicle's charging curve and the charger's power; at DC stations with \texttt{shared\_power} each car gets less power when the station is busy. Charging prices can vary by time of day (\texttt{price.by\_time}; the price when delivery starts applies). Some chargers can be out of service, and a charger can fail during charging. A charged car left connected pays \texttt{idle\_fee} after \texttt{grace\_min}.
\item charge and refuel accept \texttt{max\_wait\_min}: if the queue is longer, nothing is delivered.
\item Unattended charging can run while the person does something else, as a parallel block of two lanes, when the charger allows it (\texttt{unattended\_allowed}).
\end{itemize}
\par\smallskip\textbf{Parking}\par
\begin{itemize}[leftmargin=1.1em,itemsep=1pt,topsep=1pt,parsep=0pt]
\item Car parks and energy services have \texttt{total\_places} shared with other drivers; they fill and free during the day. If a car park is full, park is rejected and you are told how many places are taken and how many cars queue. \texttt{queue\_for\_space} with \texttt{max\_wait\_min} waits at the entrance: you get a place if one frees up in time, otherwise you are still at the entrance after \texttt{max\_wait\_min}. Or go elsewhere.
\item Prices are in each service's price: \texttt{per\_hour} with \texttt{free\_min} and a daily cap; tiers give a dearer first hour; \texttt{validation\_min} is free time when the person's activity at the venue the car park belongs to is done during the stay; \texttt{max\_stay\_min} limits the stay and time beyond it costs \texttt{overtime\_factor} times as much, outside the cap. Some car parks and fuel stations close at night (\texttt{open\_interval\_min}).
\end{itemize}
\par\smallskip\textbf{Activities and the people in the car}\par
\begin{itemize}[leftmargin=1.1em,itemsep=1pt,topsep=1pt,parsep=0pt]
\item Venues keep opening hours (\texttt{open\_min}): arriving early means waiting at the door, and the activity must end by closing time.
\item An activity with \texttt{start\_min} begins then (an appointment, a screening, a pick-up, a booking). Arriving early means waiting until it starts; arriving more than \texttt{late\_grace\_min} after \texttt{start\_min} means it is lost for today (rejected).
\item An activity with \texttt{location\_options} names a kind of place; doing it at any candidate satisfies it.
\item \texttt{conditions.companion} (an elderly person or a child) is with the user and walks at most \texttt{conditions.companion\_walk\_limit\_m} on one walk (less in rain or snow, already in that figure); a longer walk is rejected. At a venue's entrance you may \texttt{drop\_off} the companion (they wait at the door) and later \texttt{pick\_up} them there; they must be in the car to go elsewhere or home. \texttt{pick\_up} and \texttt{drop\_off} are only for this companion; a friend the user asks you to collect is an ordinary stop.
\item Steps a person does without thinking are done for you and recorded: getting in, leaving a car park (paying) when driving off, walking between the car and the venue, parking in the charging bay before charging, and parking in the venue's own car park when an activity is started at its entrance (rejected, with the reading, if it is full).
\end{itemize}
\par\smallskip\textbf{On the way}\par
\begin{itemize}[leftmargin=1.1em,itemsep=1pt,topsep=1pt,parsep=0pt]
\item Driving, charging, queueing and activities can pause when something matters to the user: an incident ahead, a risk of being late, energy that will not last, the car park ahead filling up (live-data car parks), a price change, a charger fault, queue progress, or the user speaking. Waiting stops when the user speaks. The result then carries interruption and the action is paused: continue (for charging, optionally with a new amount), \texttt{stop\_charge}, or do something else. A paused drive is at a junction: you may continue or drive another route from it (\texttt{query\_routes} with origin=\texttt{\textless{}junction id\textgreater{}}); you cannot park, walk or wait there.
\item The user may add a stop, ask to be home earlier, cancel an activity or move an appointment; the result shows the message and the task changes from then on.
\item Every result carries navigation: if the car left now by the quickest route as traffic is known now, when each unfinished activity and home would be reached, the margin to the user's time, and the energy on arriving home.
\end{itemize}
\par\smallskip\textbf{How the day is judged}\par
\begin{itemize}[leftmargin=1.1em,itemsep=1pt,topsep=1pt,parsep=0pt]
\item The day succeeds only if all of these hold: the person, the car and any companion are home; every activity the user requires (including ones added during the trip, and not ones cancelled) is done; and the energy on arriving home is at least the energy needed for tomorrow's drive (see Energy; a shortfall of up to 10\% of \texttt{recommended\_minimum\_energy} is tolerated). Otherwise the day fails.
\item Also reported: whether set times were kept and home was reached by the user's time, how late, whether requests made during the trip were followed, and, on successful days, where the user's time went (driving, walking, finding a place and queueing, waiting for energy, waiting before a set time, other waiting, the whole trip from leaving home), how far the user walked, and the money spent on parking, energy, tolls and fees. Waiting at home before the first drive does not count as trip time.
\end{itemize}
\par\smallskip \textbf{Crashes:} some incidents close their stretch of road for a while (police and recovery at work), after which traffic crawls past until they clear. When you meet one, the drive pauses with \texttt{incident\_ahead} and says until when the road is closed. Driving on means waiting in the queue until it reopens. A drive you start on a route through a closure you know of is rejected and tells you when it reopens: take another route or leave later.\par
\par\smallskip \textbf{Night rates:} some car parks charge a lower hourly rate at night; their price then has night (\texttt{from\_min}, \texttt{to\_min}, \texttt{per\_hour}), and that rate applies to the parking time inside those hours. A DC fast charger stops at 90\% of capacity: ask a DC charger for at most that, and use an AC charger to go above it. When a car park is full and drivers are turned away, some of them try the car parks nearby, which then fill sooner.\par
\par\smallskip \textbf{Notices:} \texttt{get\_task} may list notices, such as an event near a place of the trip with its hours; the roads around it are busy before it starts and after it ends, and car parks near it fill up before the start. Route forecasts are those of an ordinary day and do not include it; live readings do. Weather: \texttt{conditions.weather\_forecast} (also said in the request) may give a change of weather later in the day with its probability; rain or snow slows driving and walking and halves how far a frail companion can walk. Reservations: a car park or charging station whose price has reservation (fee, \texttt{hold\_min}) can be booked with \texttt{\{"type":"reserve",\allowbreak{}"service\_id":...,\allowbreak{}"arrive\_at\_min":...\}}: the fee is paid now, and arriving from 15 min before that time until \texttt{hold\_min} after it, the place is ready (no queue); after that the booking lapses. Appointments: for an activity with a set time you can phone the people waiting with \texttt{\{"type":"contact",\allowbreak{}"activity\_id":...\}}; the result (agreed or declined) is in the record; when they agree they wait 20 min longer before the appointment is lost. Arriving late still counts as late. A reservation or a call takes no time.\par
\par\smallskip \textbf{Compact responses:} \texttt{execute\_action} returns the records added by that action (\texttt{completed\_since\_last\_action}), not the whole history, and what the user just said with what it changes (\texttt{new\_user\_messages}); \texttt{observe\_state} returns the whole history (\texttt{completed\_prefix}) and all of the user's changes (\texttt{task\_updates}). \texttt{query\_routes} gives one row per route in the order of \texttt{route\_columns}. \texttt{search\_services} gives each facility's hourly forecast as \texttt{forecast\_rows}, in the order of \texttt{forecast\_columns}, from the current hour on. Numbers are rounded to two decimals. Query tools can be called several in the same turn; \texttt{execute\_action} still goes alone, and you read its result before acting again.\par
\end{tcolorbox}

\end{document}